\documentclass[journal]{IEEEtran}
\ifCLASSOPTIONcompsoc
  \usepackage[nocompress]{cite}
\else
  \usepackage{cite}
\fi
\ifCLASSINFOpdf
    \usepackage[pdftex]{graphicx}
\else
   \usepackage[dvips]{graphicx}
\fi
\usepackage{amsmath}
\usepackage{bm}
\usepackage{makecell}
\usepackage{array}

\ifCLASSOPTIONcompsoc
  \usepackage[caption=false,font=footnotesize,labelfont=sf,textfont=sf]{subfig}
\else
  \usepackage[caption=false,font=footnotesize]{subfig}
\fi
\usepackage{threeparttable}

\usepackage{fixltx2e}

\usepackage{stfloats}

\usepackage{subfloat}
\usepackage{url}
\usepackage{booktabs}

\usepackage{amsfonts}

\begin{document}
%
\title{Accurate Monocular Visual-inertial SLAM using a Map-assisted EKF Approach}
%
%
%
%

\author{Meixiang~Quan,
        Songhao~Piao,
        Minglang~Tan,
        Shi-Sheng~Huang
\IEEEcompsocitemizethanks{\IEEEcompsocthanksitem S. Piao(piaosh@hit.edu.cn) and S. Huang(shishenghuang.net@gmail.com) are the corresponding authors}}
\maketitle

\begin{abstract}
This paper presents a novel tightly-coupled monocular visual-inertial Simultaneous Localization and Mapping algorithm, which provides accurate and robust localization within the globally consistent map in real time on a standard CPU. This is achieved by firstly performing the visual-inertial extended kalman filter(EKF) to provide motion estimate at a high rate. However the filter becomes inconsistent due to the well known linearization issues. So we perform a keyframe-based visual-inertial bundle adjustment to improve the consistency and accuracy of the system. In addition, a loop closure detection and correction module is also added to eliminate the accumulated drift when revisiting an area. Finally, the optimized motion estimates and map are fed back to the EKF-based visual-inertial odometry module, thus the inconsistency and estimation error of the EKF estimator are reduced. In this way, the system can continuously provide reliable motion estimates for the long-term operation. The performance of the algorithm is validated on public datasets and real-world experiments, which proves the superiority of the proposed algorithm.
\end{abstract}


\IEEEdisplaynontitleabstractindextext

%
\IEEEpeerreviewmaketitle

\vspace{0.5in}

\IEEEraisesectionheading{\section{Introduction}\label{sec:introduction}}
Concurrent motion estimation and map reconstruction by combining visual and inertial measurements has received significant interest in the field of Robotics and Computer Vision communities. This visual-inertial sensor suite can serve as an ideal alternative to GPS in environments where GPS is denied, since both sensors are small, lightweight, cheap enough and complementary. On the one hand, Visual SLAM can provide good tracking and rich map information in visually distinguishable environments. However due to the sensor limitation, visual simultaneous localization and mapping(SLAM) is sensitive to motion blur, occlusions and illumination changes. On the other hand, inertial sensors are able to provide self-motion information at high frequency, so inertial navigation is robust to aggressive motion and can provide absolute scale for the motion. Whereas the result of inertial navigation is noisy and diverges even in a few seconds. Therefore, fusing measurements from the inertial sensor to the visual SLAM in a tightly-coupled way, both the robustness and the accuracy of motion tracking can be dramatically improved. The advantage of visual-inertial sensor fusion is the most obvious in a monocular visual-inertial setup, because the scale of the motion estimation and map structure computed from monocular SLAM is ambiguous and liable to drift over time.

In this paper, we aim to build a system which enables the accurate and robust motion tracking within the accurately and consistently reconstructed map. Firstly, we employ the EKF-based visual-inertial odometry(VIO) to track the 3D motion of the IMU body frame in real time. Whereas due to the linearization errors, the estimator tends to be inconsistent, which results in large estimation errors and divergence. So we intend to perform bundle adjustment(BA) to reduce the inconsistency of the estimator. It is demonstrated in \cite{WHYFILTER} that the BA techniques can achieve better accuracy than filtering techniques, because optimization methods can iteratively relinearize measurement equations to better deal with their nonlinearity. However, the optimization method leads to a high computational cost. In addition, increasing the number of feature correspondences and keyframes in the window of local BA will lead to significant increase in accuracy, whereas feature extraction and matching, optimization for the local BA are also time consuming. So if we increase the number of features, and perform the local BA after the EKF for each frame to increase the accuracy of the system, the system will incapable of real time operation. Thus, in order to ensure both real time operation and accurate global map reconstruction, we extract a vast number of features only for those selected keyframes, we use these features to construct global map and perform the local BA in a parallel thread. In this way, we can properly solve the loss of accuracy due to the linearization in real-time. Besides, if the system is unable to close loops, the drift of the estimated trajectory will accumulate without bound, even if the sensor is continuously revisiting the same place. Therefore, we add a loop closure module in a new parallel thread for reducing the accumulated drift when returning to an already mapped area.

In summary, we propose a tightly-coupled monocular visual-inertial SLAM(VISLAM) system, which is able to perform real-time, accurate, robust and long term localization and map reconstruction. Our approach operates in three parallel threads, one thread is used to perform the EKF VIO, and the other two threads, one for BA and the other one for loop closing, are used to compensate for the error growth of the estimated trajectory and construct an accurate and consistent global map. In VIO thread, since the computational cost of the EKF is quadratic in the number of features, for the real-time state estimation we extract the appropriate number of features for each frame, and extract a lot of features only in selected keyframes for performing local BA and loop closure. Finally, we design a feedback mechanism to increase both the consistency and accuracy of the EKF estimator, which is achieved by feeding back the optimized state and globally consistent map to update the state vector of the EKF VIO module.

In experiments, the results demonstrate the benefits of our system towards the EKF based VIO. We also compare to the state-of-the-art VIO and VISLAM approaches, and demonstrate the superior performance of our method.

\begin{figure*}[htbp]
\centering
\includegraphics[width=5.5in]{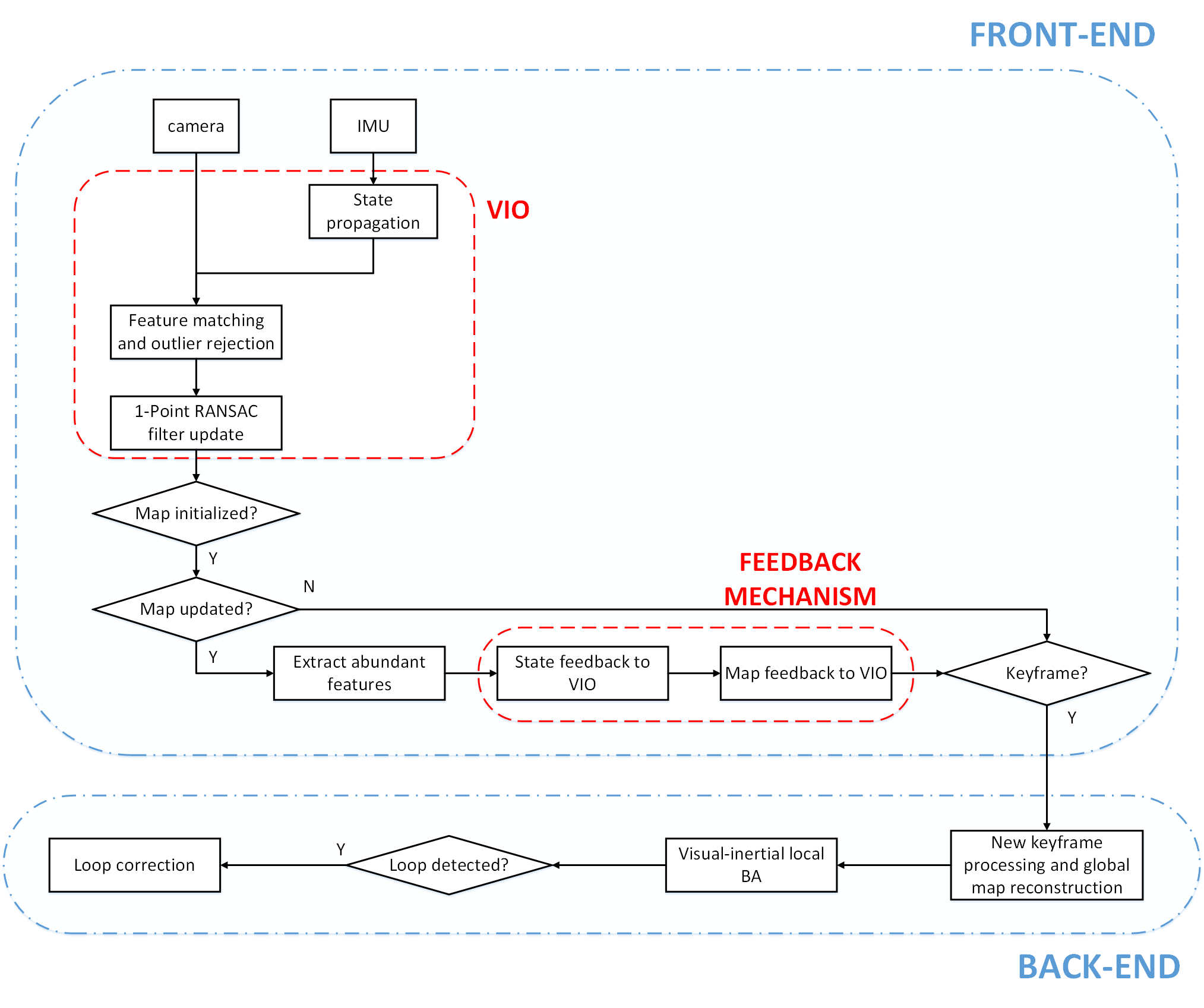}
\caption{An overview of the proposed monocular VISLAM algorithm, which contains two main components: EKF based VIO front-end and nonlinear optimization and loop closure back-end.}
\label{fig_chart}
\end{figure*}

Our monocular visual-inertial SLAM algorithm is shown in Fig.\ref{fig_chart}. The system is complete and drift-free in large scale environments. The remainder of the paper is organized as follows. In Section \ref{relatedwork}, we describe the relevant literature. Notations are given in Section \ref{PRELIMINARIES}. In Section \ref{sec:filter}, EKF based VIO is presented and the Jacobian matrices of EKF are given in appendix. A tightly-coupled joint visual-inertial nonlinear optimization is introduced In Section \ref{sec:optimization}. Section \ref{sec:approach} introduce our tightly-coupled VISLAM approach. Experiments results are shown in Section \ref{experiment}. Finally, the paper is concluded in Section \ref{conclusion}.

\section{RELATED WORK} \label{relatedwork}

There are a vast amount of research towards visual SLAM problem, we refer to the review paper \cite{Cadena2016Past} for the progress made in the past few decades. In this section, we will discuss the most relevant works on monocular VIO and VISLAM system.

The fusion for visual and inertial measurements is usually divided into two classes. Loosely-coupled approaches, e.g. \cite{LOOSE1}\cite{LOOSE2}, process the visual and inertial measurements separately. Therefore the accumulated drift in vision module cannot be eliminated from the usability of inertial measurements, which leads the resulted estimate to be sub-optimal. Tightly-coupled ones interested in this work, e.g. \cite{VISLAMINVDEP, VISLAMGPSDENIED, VINAVIGATION, FEJ, OCEKF, MSCKF, MSCKFCON1, MSCKF2, MSCKFCON2,OKVIS,GTSAM,DIRECTVI,ORBVIO,LSDVIO,VINSMONO1}, perform VIO or VISLAM system by considering the tight interaction between visual and inertial measurements to optimally exploit the both sensing cues, thus achieve higher precision at the expensive of additional computational complexity. Besides, for tightly-coupled VIO/VISLAM solutions, two methodologies have been prevalent: filtering-based methods \cite{VISLAMINVDEP, VISLAMGPSDENIED, VINAVIGATION, FEJ, OCEKF, MSCKF, MSCKFCON1, MSCKF2, MSCKFCON2} and BA-based methods \cite{OKVIS,GTSAM,DIRECTVI,ORBVIO,LSDVIO,VINSMONO1,VINSMONO2}.

Historically, the monocular SLAM problem has been addressed with filtering method, which operates on the mean and covariance of the probabilistic distribution in a kalman filtering framework. Filtering based approaches require fewer computational resources due to the continuous marginalization of past state, however the system get slightly lower accuracy due to the linearization error. According to the way processing the measurement information, the recursive filtering approaches can be classified into two main categories: extended kalman filter (EKF) based methods \cite{VISLAMINVDEP, VISLAMGPSDENIED, VINAVIGATION, FEJ, OCEKF} and sliding window filtering approaches \cite{MSCKF, MSCKFCON1, MSCKF2, MSCKFCON2}. The state vector of EKF-based SLAM algorithms include both the pose of the platform and a set of feature positions, so as long as these features are continuously observed and contained in the state vector, the estimated pose relative to these features will not drift. However it have high computational complexity (quadratic in the number of features in the state vector), therefore only currently observable landmarks are tracked to ensure real-time operation. In contrast, sliding window filtering approaches maintain a sliding window of past camera poses in the state vector, and use the feature measurements to impose probabilistic constraints on these poses, therefore keep computational complexity only linear in the number of features by excluding point features from filter state vector. Generally, the VIO problem has four unobservable directions, but since the linearization errors, the system only have three unobservable directions, which renders the filter inconsistent. So papers \cite{FEJ, OCEKF, MSCKFCON1, MSCKF2, MSCKFCON2} were proposed a series of methods, e.g. first-estimates Jacobian and constraint of system observability, to improve the consistency of the system. If the measurement models were linear, both methods yield the same result equal to the MAP estimate.

In \cite{WHYFILTER}, it was shown that nonlinear optimization-based approaches provide better accuracy than filtering approaches by its capability to relinearize the state at each iteration, therefore avoiding integrated error from linearization, however it leads to higher computational demands. In following, we introduce several classic tightly-coupled BA-based VIO and VISLAM system. OKVIS \cite{OKVIS} presented an approach to tightly integrate inertial measurements into keyframe-based visual SLAM, which makes the nonlinear cost function comprised of IMU error term with the landmark reprojection error term to be jointly optimized. Additionally, marginalization of old state is used to maintain a bounded-sized optimization window, therefore OKVIS has achieved increased accuracy and robustness in real-time operation. However, the system needs to repeatedly compute the IMU integration when the linearization point changes. To eliminate this repeated computation, Foster et al. provided a preintegration theory for inertial measurements in \cite{GTSAM} that properly address the manifold structure of the rotation group based on the work of \cite{VIA}. Then, the preintegrated IMU model and structureless visual model are seamlessly integrated in a fully probabilistic manner to build a much more computationally efficient optimization method for state estimation. Therefore, the system achieves better accuracy than Project Tango \cite{TANGO} by using SVO as front-end and the visual-inertial joint optimization in back-end. Tightly-coupled visual-inertial odometry methods mentioned above all lack the capability to close loops and reuse an already reconstructed map, due to the marginalization of past states to maintain a constant computational cost or the use of full smoothing. Thus, ORB-VISLAM \cite{ORBVIO} presented a real-time tightly-coupled monocular visual-inertial SLAM system, which enables the loop closure and the reuse of previously estimated 3D map. The system achieved higher accuracy than the fully direct, stereo visual-inertial odometry \cite{LSDVIO}, because there is no drift accumulation for localization in already mapped areas. Recently, a novel real-time, tightly-coupled, sliding window optimization based versatile monocular visual-inertial odometry \cite{VINSMONO1}\cite{VINSMONO2} was proposed, in which, the state of the system and a representation of the environment are estimated by local BA in one thread, and loops are closed in lightweight manner in parallel thread.

Both filtering-based methods and BA-based methods have their merits, so in this work, we tightly fuse both methods to achieve the best accuracy, robustness and efficiency.

\section{NOTATIONS}\label{PRELIMINARIES}
Throughout the paper, we denote the world reference frame as $(\cdot)^W$, and denote the IMU body frame and camera frame for the $k^{th}$ image as $(\cdot)^{B_k}$ and $(\cdot)^{C_k}$ respectively. In addition, we employ $\bm{\mathrm{R}}^{\mathcal{F}_1}_{\mathcal{F}_2} \in \bm{\mathrm{SO}}(3)$ to represent rotation from frame $\{\mathcal{F}_2\}$ to $\{\mathcal{F}_1\}$, ${\bm{\mathrm{p}}}^{\mathcal{F}_1}_{\mathcal{F}_2} \in \mathbb{R}^3$ and ${\bm{\mathrm{v}}}^{\mathcal{F}_1}_{\mathcal{F}_2} \in \mathbb{R}^3$ to describe the 3D position and velocity of frame $\{\mathcal{F}_2\}$ with respect to the frame $\{\mathcal{F}_1\}$. Besides, the rotation and translation between the rigidly mounted camera-IMU sensor are denoted as $\bm{\mathrm{R}}^B_C$ and $\bm{\mathrm{p}}^B_C$, which was computed from the calibration.

For the over-parameterized rotation matrix, a vector $\bm{\mathrm{\xi}} \in \mathbb{R}^3$ can be computed from the tangent space $\mathfrak{so}(3)$ of manifold $\bm{\mathrm{SO}}(3)$ to provide a minimal representation. The Lie algebra $\mathfrak{so}(3)$ and Lie group $\bm{\mathrm{SO}}(3)$ are related through the logarithm map and exponential map:
\begin{equation}
\bm{\xi} = \mathrm{Log}(\bm{\mathrm{R}}) = {\mathrm{log}(\bm{\mathrm{R}})}^{\vee}
\end{equation}
\begin{equation}
\bm{\mathrm{R}}(\bm{\xi}) = \mathrm{Exp}(\bm{\xi}) = \mathrm{exp}({\bm{\xi}}^{\wedge})
\end{equation}
where ${(\cdot)}^{\wedge}$ operator maps a vector in $\mathbb{R}^3$ to a $3 \times 3$ skew symmetric matrix, and ${(\cdot)}^{\vee}$ is the inverse operator. The formula for $\mathrm{log}(\cdot)$ and $\mathrm{exp}(\cdot)$ can be found in \cite{Stochastic}.

Furthermore, the uncertainty of rotation $\bm{\mathrm{R}} \in \bm{\mathrm{SO}}(3)$ and $\bm{\mathrm{\xi}} \in \mathbb{R}^3$ are described as:
\begin{equation}
\bm{\mathrm{R}} = \hat{\bm{\mathrm{R}}} \oplus \delta \bm{\xi} =\hat{\bm{\mathrm{R}}} \mathrm{Exp}(\delta \bm{\xi})
\end{equation}
\begin{equation}
\bm{\mathrm{\xi}} = \hat{\bm{\mathrm{\xi}}} \oplus \delta \bm{\xi} = \mathrm{Log}(\mathrm{Exp}(\hat{\bm{\mathrm{\xi}}}) \mathrm{Exp}(\delta \bm{\xi}))
\label{rotationerror}
\end{equation}
where $\hat{\bm{\mathrm{R}}}$ and $\hat{\bm{\mathrm{\xi}}}$ are the mean estimate of $\bm{\mathrm{R}}$ and $\bm{\mathrm{\xi}}$ respectively, and $\delta \bm{\xi} \sim \mathcal{N}(\bm{0},\bm{\mathrm{\Sigma}})$ is a normally distributed perturbation.

For visual measurements, we consider a projection function $\pi: \mathbb{R}^3 \to \Omega \subset \mathbb{R}^2$, which projects the $l^{th}$ map point expressed in the current camera frame $ \bm{\mathrm{f}}^{C_k}_l = {[x^{C_k}_l \ y^{C_k}_l \ z^{C_k}_l]}^\mathrm{T} \in \mathbb{R}^3$ onto 2D points on the image plane $\bm{\mathrm{z}}_{kl} = {[u_{kl} \ v_{kl}]}^\mathrm{T} \in \Omega$:
\begin{equation}
\begin{split}
\widetilde{\bm{\mathrm{z}}}_{kl} &= \bm{\mathrm{z}}_{kl} + \bm{\sigma}_{kl}\\
&=\pi(\bm{\mathrm{f}}^{C_k}_l) + \bm{\sigma}_{kl}\\
&=\left[ \begin{array}{c} f_x \frac{x^{C_k}_l}{z^{C_k}_l} + c_x \\ f_y \frac{y^{C_k}_l}{z^{C_k}_l} + c_y \end{array} \right] + \bm{\sigma}_{kl}\\
\label{visualmm}
\end{split}
\end{equation}
where $\widetilde{\bm{\mathrm{z}}}_{kl}$ is the corresponding feature measurement, and $\bm{\sigma}_{kl}$ is the $2 \times 1$ measurement noise with covariance $\bm{\Sigma}_{\sigma_{kl}}$ associated to the feature scale. In addition, $f_x$, $f_y$ are focal length and $c_x$, $c_y$ are principle point, which are known from calibration.


\section{VIO Description}\label{sec:filter}
In this section, we describe the EKF VIO system, which is based on the work of \cite{MONOSLAM}. An overview of the algorithm is given in VIO part of Fig. \ref{fig_chart}. Inertial measurements are used to predict the motion movement in the prediction stage, then the state is updated using the matched visual features. In this way, we can estimate the state of the body frame efficiently.

\subsection{Full State Vector}
The state vector to be estimated comprises the IMU state $\bm{\mathrm{X}}_{B_k}$ and a set of landmark parameters ${\bm{\mathrm{X}}_{L_k}}$:
\begin{equation}
\bm{\mathrm{X}}_k = {[\bm{\mathrm{X}}_{B_{k}}^\mathrm{T} \ \bm{\mathrm{X}}_{L_k}^\mathrm{T}]}^\mathrm{T}
\end{equation}

The IMU state is formulated by the vector:
\begin{equation}
\bm{\mathrm{X}}_{B_{k}} = {[ {\bm{\mathrm{\xi}}^{W}_{B_k}}^\mathrm{T} \ {\bm{\mathrm{p}}^{W}_{B_k}}^\mathrm{T} \ {\bm{\mathrm{v}}^{W}_{B_k}}^\mathrm{T} \ {\bm{\mathrm{b}}_{a_k}^\mathrm{T}} \ {\bm{\mathrm{b}}_{g_k}^\mathrm{T}}]}^\mathrm{T}
\label{imustate}
\end{equation}
where ${\bm{\mathrm{\xi}}^{W}_{B_k}} \in \mathbb{R}^3$ is the Lie algebra of orientation ${\bm{\mathrm{R}}^{W}_{B_k}} \in \bm{\mathrm{SO}}(3)$ from frame $\{B_k\}$ to $\{W\}$, ${\bm{\mathrm{p}}^{W}_{B_k}} \in \mathbb{R}^3$ and ${\bm{\mathrm{v}}^{W}_{B_k}} \in \mathbb{R}^3$ are the 3D position and velocity of frame $\{B_k\}$ with respect to $\{W\}$, as well as $\bm{\mathrm{b}}_a \in \mathbb{R}^3$ and $\bm{\mathrm{b}}_g \in \mathbb{R}^3$ are additive accelerometer and gyroscope biases respectively. Following \eqref{imustate}, the IMU error state vector is defined as:
\begin{equation}
\delta \bm{\mathrm{X}}_{B_{k}} = {[ \delta {\bm{\mathrm{\xi}}_{B_k}^\mathrm{T}} \ \delta {\bm{\mathrm{p}}^{W}_{B_k}}^\mathrm{T} \ \delta {\bm{\mathrm{v}}^{W}_{B_k}}^\mathrm{T} \ \delta {\bm{\mathrm{b}}_{a_k}^\mathrm{T}} \ \delta {\bm{\mathrm{b}}_{g_k}^\mathrm{T}}]}^\mathrm{T}
\end{equation}
where we use the standard additive error for the 3D position, velocity and biases, while for rotation, the error is defined as \eqref{rotationerror}.

Assuming that $m$ features are included in the map at time-step $k$, then the coordinates of features are:
\begin{equation}
\bm{\mathrm{X}}_{L_k} = {[{{\bm{\mathrm{f}}_1^W}^\mathrm{T}} \ \cdots \ {{\bm{\mathrm{f}}_m^W}^\mathrm{T}}]}^\mathrm{T}
\end{equation}

The position of the $l^{th}$ landmark ${\bm{\mathrm{f}}_l^W}$ is paramterized in inverse depth coordinates as:
\begin{equation}
{\bm{\mathrm{f}}_l^W} = {[ x_l \ y_l \ z_l \ \theta_l \ \phi_l \ \rho_l ]}^\mathrm{T}
\end{equation}
where ${(x_l, y_l, z_l)}^\mathrm{T}$ is the camera position, in which the $l^{th}$ landmark was firstly observed. $\theta_l$ and $\phi_l$ are the azimuth and elevation angle defining unit ray(expressed in the world frame) that goes from the camera center ${(x_l, y_l, z_l)}^\mathrm{T}$ to the $l^{th}$ landmark, and ${\rho _l}$ is its inverse depth along the unit ray.

Therefore the EKF error state vector is expressed as:
\begin{equation}
\delta \bm{\mathrm{X}}_k = {[ \delta \bm{\mathrm{X}}_{B_{k}}^\mathrm{T} \ \delta {{\bm{\mathrm{f}}_1^W}^\mathrm{T}} \ \cdots \ \delta {{\bm{\mathrm{f}}_m^W}^\mathrm{T}}]}^\mathrm{T}
\end{equation}
where we use the standard additive error for the landmark.

\subsection{IMU Propagation Model}
Different to many other visual-inertial methods, we define the state propagation model directly in discrete-time to make the derivatives required for the EKF prediction are available in close-form. Using the measured acceleration $\widetilde{\bm{\mathrm{a}}}_{k-1}$ and angular velocity $\widetilde{\bm{\mathrm{\omega}}}_{k-1}$ obtained from IMU, the discrete-time propagation model $\bm{\mathrm{X}}_{B_{k|k-1}} = \bm{\mathrm{f}}_{k}(\bm{\mathrm{X}}_{B_{k-1}})$ is:
\begin{equation}
\begin{split}
&{\bm{\mathrm{\xi}}^{W}_{B_{k|k-1}}} = \mathrm{Log} \left(\mathrm{Exp}({\bm{\mathrm{\xi}}^{W}_{B_{k-1}}}) \mathrm{Exp}\left((\widetilde{\bm{\mathrm{\omega}}}_{k-1}-\bm{\mathrm{b}}_{g_{k-1}}-\bm{\mathrm{n}}_{gd}) \Delta t \right)\right)\\
&{\bm{\mathrm{p}}^{W}_{B_{k|k-1}}} = {\bm{\mathrm{p}}^{W}_{B_{k-1}}} + {\bm{\mathrm{v}}^{W}_{B_{k-1}}} \Delta t\\
&{\bm{\mathrm{v}}^{W}_{B_{k|k-1}}} = {\bm{\mathrm{v}}^{W}_{B_{k-1}}} + ({{\bm{\mathrm{R}}^{W}_{B_{k-1}}}(\widetilde{\bm{\mathrm{a}}}_{k-1}-\bm{\mathrm{b}}_{a_{k-1}}-\bm{\mathrm{n}}_{ad})+\bm{\mathrm{g}}^W) \Delta t}\\
&{\bm{\mathrm{b}}_{a_{k|k-1}}} = {\bm{\mathrm{b}}_{a_{{k-1}}}}\\
&{\bm{\mathrm{b}}_{g_{k|k-1}}} = {\bm{\mathrm{b}}_{g_{{k-1}}}}
\end{split}
\end{equation}
where $\bm{\mathrm{R}}^{W}_{B_{k-1}} = \mathrm{Exp}({\bm{\mathrm{\xi}}^{W}_{B_{k-1}}})$, $\bm{\mathrm{n}}_{gd} \sim \mathcal{N}(\bm{\mathrm{0}}, {\bm{\Sigma}_{g}}/{\Delta t})$ and $\bm{\mathrm{n}}_{ad} \sim \mathcal{N}(\bm{\mathrm{0}},\bm{\Sigma}_{a}/\Delta t)$ are discrete-time white Gaussian noise for inertial measurements, and $\bm{\mathrm{g}}^W$ is the constant gravity. In this work, we ignore the slow random walk of the inertial biases, so the biases are considered fixed and estimated as part of the state. The linearized discrete-time IMU error state propagation model is represented as:
\begin{equation}
\delta \bm{\mathrm{X}}_{B_{k|k-1}} = \bm{\mathrm{\Phi}}_{k} \delta \bm{\mathrm{X}}_{B_{k-1}} + \bm{\mathrm{G}}_{k} \bm{\mathrm{n}}_{B}
\label{errorstatetrasition}
\end{equation}
where $\bm{\mathrm{n}}_{B} = {[{\bm{\mathrm{n}}_{ad}}^\mathrm{T} \ \ {\bm{\mathrm{n}}_{gd}}^\mathrm{T}]}^\mathrm{T}$ is the system noise with covariance $\bm{\mathcal{Q}} = \left[ \begin{array} {cc} \bm{\Sigma}_{a}/\Delta t & \bm{\mathrm{0}}_{3 \times 3} \\ \bm{\mathrm{0}}_{3 \times 3} & \bm{\Sigma}_{g}/\Delta t   \end{array}\right]$. The matrices $\bm{\mathrm{\Phi}}_{k}$ and $\bm{\mathrm{G}}_{k}$ in \eqref{errorstatetrasition} are Jacobians of $\bm{\mathrm{f}}_k(\cdot)$ with respect to IMU state and the system noise, these representation can be found in appendix A.

Therefore, the covariance matrix is propagated as follows:
\begin{equation}
\begin{split}
\bm{\mathrm{P}}_{k|k-1} &= \left[ \begin{array}{cc} \bm{\mathrm{P}}_{B_{k|k-1}} & \bm{\mathrm{P}}_{BL_{k|k-1}}  \\
\bm{\mathrm{P}}_{LB_{k|k-1}} & \bm{\mathrm{P}}_{L_{k|k-1}} \end{array}\right]  \\
&= \left[ \begin{array}{cc} \bm{\mathrm{\Phi}}_{k} \bm{\mathrm{P}}_{B_{k-1}}{\bm{\mathrm{\Phi}}_{k}}^\mathrm{T}+\bm{\mathrm{G}}_{k} \bm{\mathcal{Q}}{\bm{\mathrm{G}}_{k}}^\mathrm{T} & \bm{\mathrm{\Phi}}_{k} \bm{\mathrm{P}}_{BL_{{k-1}}}  \\
\bm{\mathrm{P}}_{LB_{{k-1}}} {\bm{\mathrm{\Phi}}_{k}}^T & \bm{\mathrm{P}}_{L_{{k-1}}} \end{array}\right]  \\
\end{split}
\end{equation}

\subsection{Measurement Model}
The inverse depth parameterization is used for features to (1) enhance the degree of linearity for measurement equations, and (2) better deal with the low parallax features. The Euclidean XYZ coordinates of the $l^{th}$ feature in world frame $\bm{\mathrm{y}}_l^W$ can be transformed from its inverse depth representation $\bm{\mathrm{f}}_l^W$ as:
\begin{equation}
\bm{\mathrm{y}}_l^W =  \left[ \begin{array}{c} x_l \\ y_l \\ z_l \end{array}\right] + \frac{1}{\rho_l} \bm{\mathrm{m}}(\theta_l, \phi_l)
\end{equation}
\begin{equation}
\bm{\mathrm{m}}(\theta_l, \phi_l) = \left[ \begin{array}{c}
\mathrm{cos}\phi_l \mathrm{sin}\theta_l \\ -\mathrm{sin}\phi_l \\ \mathrm{cos}\phi_l \mathrm{cos}\theta_l \end{array}\right]
\end{equation}

Thus the measurement model representing the projection of the $l^{th}$ landmark to the ${k}^{th}$ image is:
\begin{equation}
\begin{split}
&\widetilde{\bm{\mathrm{z}}}_{kl} = \bm{\mathrm{h}}_{kl}(\bm{\mathrm{X}}_{B_{k|k-1}},\bm{\mathrm{f}}_l^W) + \bm{\mathrm{\sigma}}_{kl}\\
&=\pi \left( \bm{\mathrm{f}}_l^{C_k} \right) +\bm{\mathrm{\sigma}}_{kl}
\end{split}
\end{equation}
where $\bm{\mathrm{f}}_l^{C_k} = \bm{\mathrm{R}}_{W}^{C_{k|k-1}} \left(\rho_l \left(\left[ \begin{array}{c} x_l \\ y_l \\ z_l \end{array}\right]-\bm{\mathrm{p}}^{W}_{C_{k|k-1}}\right) + \bm{\mathrm{m}}(\theta_l, \phi_l) \right)$, $\bm{\mathrm{R}}_{W}^{C_{k|k-1}} = {(\bm{\mathrm{R}}^{W}_{B_{k|k-1}} \bm{\mathrm{R}}^{B}_{C})}^\mathrm{T}$ and $\bm{\mathrm{p}}^{W}_{C_{k|k-1}} = {\bm{\mathrm{p}}^{W}_{B_{k|k-1}}} + \bm{\mathrm{R}}^{W}_{B_{k|k-1}} {\bm{\mathrm{p}}^{B}_{C}}$. From the measurement model, we compute the reprojection error and its linearized approximation as:
\begin{equation}
\begin{split}
\bm{\mathrm{r}}_{kl} &= \widetilde{\bm{\mathrm{z}}}_{kl} - \bm{\mathrm{h}}_{kl}({\bm{\mathrm{X}}}_{B_{k|k-1}}, {({\bm{\mathrm{X}}}_{L_{k}})}_l)\\
& \simeq \bm{\mathrm{H}}_{B_{kl}} \delta \bm{\mathrm{X}}_{B_{k|k-1}} + \bm{\mathrm{H}}_{f_{kl}} \delta \bm{\mathrm{f}}_l^W + \bm{\mathrm{\sigma}}_{kl}
\label{ekfmeasurementmodel}
\end{split}
\end{equation}
where the matrices $\bm{\mathrm{H}}_{B_{kl}}$ and $\bm{\mathrm{H}}_{f_{kl}}$ are derivatives of the measurement model with respect to the IMU state estimate and the $l^{th}$ feature position respectively, their expressions are given in appendix B.

Therefore, we can obtain the measurement Jacobian matrix as:
\begin{equation}
\bm{\mathrm{H}}_{kl} = [\bm{\mathrm{H}}_{B_{kl}} \ \bm{0} \ \cdots \ \bm{\mathrm{H}}_{f_{kl}} \ \bm{0} \ \cdots]
\end{equation}

\subsection{Filter Update}
To perform an update of the estimated state, we stack the $m$ individual measurement residual $\bm{\mathrm{r}}_{kl}$ at time-step $k$ together to form a single $2m \times 1$ residual vector $\bm{\mathrm{r}}_k = {\left[ \bm{\mathrm{r}}_{k1}^\mathrm{T} \cdots \bm{\mathrm{r}}_{kl}^\mathrm{T} \cdots \bm{\mathrm{r}}_{km}^\mathrm{T} \right]}^\mathrm{T}$. In the same way, the measurement Jacobians are also combined to a single $2m \times n$ measurement matrix $\bm{\mathrm{H}}_{k} = {\left[ \bm{\mathrm{H}}_{k1}^T \cdots \bm{\mathrm{H}}_{kl}^\mathrm{T} \cdots \bm{\mathrm{H}}_{km}^\mathrm{T} \right]}^\mathrm{T}$. Then the kalman gain is computed as:
\begin{equation}
\bm{\mathrm{K}}_{k} = \bm{\mathrm{P}}_{k|k-1} \bm{\mathrm{H}}_{k}^\mathrm{T} {(\bm{\mathrm{H}}_{k}\bm{\mathrm{P}}_{k|k-1} \bm{\mathrm{H}}_{k}^\mathrm{T} + \bm{\mathrm{\Sigma}})}^{-1}
\end{equation}
where $\bm{\mathrm{\Sigma}}$ is the stacked $2m \times 2m$ measurement uncertainty. Finally, the full state and covariance are updated by:
\begin{equation}
\begin{split}
\bm{\mathrm{X}}_{{k|k}} &= \bm{\mathrm{X}}_{{k|k-1}} \circ \bm{\mathrm{K}}_{k} \bm{\mathrm{r}}_k \\
\bm{\mathrm{P}}_{k|k} &= (\bm{\mathrm{I}}-\bm{\mathrm{K}}_{k}\bm{\mathrm{H}}_{k})\bm{\mathrm{P}}_{k|k-1}
\end{split}
\end{equation}
where $\circ$ operator is equal to the $\oplus$ operator in \eqref{rotationerror} for the orientation and the addition of vector for other state.

Based on the predicted IMU state, features used to update the state are searched with the optical flow method. Prior to using each feature's measurement to update, for each matched feature, the Mahalanobis distance $d=\bm{\mathrm{r}}_{kl}^\mathrm{T} {(\bm{\mathrm{H}}_{kl} \bm{\mathrm{P}}_{k|k-1} \bm{\mathrm{H}}_{kl}^\mathrm{T} + \bm{\Sigma}_{\sigma_{kl}})}^{-1} \bm{\mathrm{r}}_{kl}$ is firstly computed to reject outliers. If $d$ is smaller than a threshold given by the 95-th percentile of the $\chi^2$ distribution, the feature is accepted as an inlier, and used for the filter update. We perform the filter update by combining 1-point RANSAC method as in \cite{1point} to find reliable inliers.

\subsection{State Augmentation}
Once a new feature $l$ is needed at time-step $k$, the initial position $\bm{\mathrm{f}}_l^W = {[ \breve{x_l} \ \breve{y_l}\ \breve{z_l} \ \breve{\theta_l} \ \breve{\phi_l} \ \breve{\rho_l} ]}^\mathrm{T}$ of the new feature is computed as follows. The camera position is computed by:
\begin{equation}
{[ \breve{x_l} \ \breve{y_l}\ \breve{z_l}]}^\mathrm{T} = \bm{\mathrm{R}}^{W}_{B_{k|k}} \bm{\mathrm{p}}^{B}_{C} + \bm{\mathrm{p}}^{W}_{B_{k|k}}
\label{stateaugmentationxyz}
\end{equation}

Besides, from the observation ${[u_{kl} \ v_{kl}]}^\mathrm{T}$ of the new feature in the image, the angles $\breve{\theta_l}$ and $\breve{\phi_l}$ defining its direction are calculated as:
\begin{equation}
\left[ \begin{array}{c} \breve{\theta_l} \\ \breve{\phi_l} \end{array}\right] = \left[ \begin{array}{c} \mathrm{arctan}(x_{kl}^W, z_{kl}^W) \\ \mathrm{arctan}(-y_{kl}^W, \sqrt{{x_{kl}^W}^2 + {z_{kl}^W}^2}) \end{array}\right]
\label{stateaugmentationthetaphi}
\end{equation}
\begin{equation}
\bm{\mathrm{\tau}}_{{kl}}^W = \left[ \begin{array}{c} x_{kl}^W \\ y_{kl}^W \\ z_{kl}^W \end{array}\right] = \bm{\mathrm{R}}^{W}_{B_{k|k}} \bm{\mathrm{R}}^{B}_{C} \left[ \begin{array}{c} \frac{u_{kl}-c_x}{f_x} \\ \frac{v_{kl}-c_y}{f_y} \\ 1 \end{array}\right]
\label{stateaugmentationuv}
\end{equation}

The initial value for ${\breve \rho _i}$ and its standard deviation $\sigma_\rho$ are set as in \cite{INVERSEDEPTH}. Then the initial position for new feature is appended to the state vector, and the state covariance matrix is also augmented accordingly:
\begin{equation}
\bm{\mathrm{P}}_{k|k} \gets \bm{\mathrm{J}}  \left[ \begin{array}{cc} \bm{\mathrm{P}}_{k|k} & \bm{\mathrm{0}}_{(15+6m) \times 6} \\ \bm{\mathrm{0}}_{6 \times (15+6m)} &  \bm{\mathrm{\Sigma}}_{h\rho} \end{array} \right] {\bm{\mathrm{J}}}^\mathrm{T}
\end{equation}
where $\bm{\mathrm{\Sigma}}_{h\rho} = \left[ \begin{array}{cc} \bm{\Sigma}_{\sigma_{kl}} & \bm{\mathrm{0}}_{2 \times 1} \\ \bm{\mathrm{0}}_{1 \times 2} & {\sigma_\rho}^2 \end{array} \right]$ denotes the uncertainty of the visual measurements and the initial inverse depth, for the Jacobian $\bm{\mathrm{J}}$ we refer the reader to appendix C.

\section{VISUAL-INERTIAL BUNDLE ADJUSTMENT}\label{sec:optimization}
Once a frame processed by EKF based VIO is selected as a keyframe, we apply a nonlinear optimization in a sliding window to improve the accuracy of the estimated state. In this section, we combine the visual and inertial measurements in an unified formulation.

\subsection{Bundle Adjustment Representation}
\begin{figure}[!t]
\centering
\includegraphics[width=3.5in]{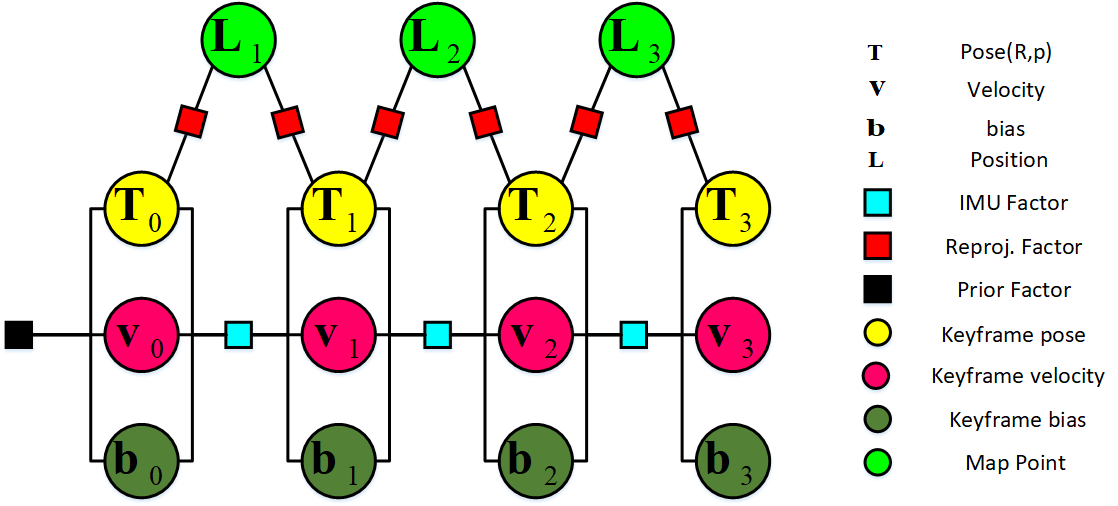}
\caption{Factor graph representing the tightly-coupled visual-inertial optimization problem in a sliding window. The states are shown as circles and factors are shown as squares. IMU factors are represented as blue squares, which is connected to the state of the previous keyframe. Red squares denote visual factors corresponding to camera observations, and black squares denote prior factors.}
\label{ba}
\end{figure}

We formulate a joint optimization problem to optimally estimate the full state in a sliding window using all the available inertial and visual measurements. Full state in sliding window contain a set of successive keyframes from $i$ to $j$ and $n$ landmarks visible by the keyframes in sliding window, which is denoted as:
\begin{equation}
    \bm{\mathcal{X}} = \{\bm{\mathrm{X}}_{B_{i}}, \ \cdots, \ \bm{\mathrm{X}}_{B_{j}}, \ \bm{\mathrm{L}}_1^W, \cdots, \ \bm{\mathrm{L}}_n^W \}
\end{equation}
where $\bm{\mathrm{L}}_l^W$ is the position of the 3D map point expressed in Euclidean XYZ coordinates. We denote the keyframes and visual measurements in sliding window as $\mathcal{K}$ and $\mathcal{C}$ respectively. Then the energy function that we want to minimize is given by:
\begin{equation}
\begin{split}
f(\bm{\mathcal{X}}) =& \ \|\bm{\mathrm{r}}_p\|^2_{{\bm{\Sigma}}_p}
+ \sum \limits_{k \in \mathcal{K}} \rho\left(\|{\bm{\mathrm{r}}_{\mathcal{I}_{k-1k}}}\|^2_{{\bm{\Sigma}}_{\mathcal{I}_{k-1k}}}\right)\\
&+ \sum \limits_{k \in \mathcal{K}, l \in \mathcal{C}} \rho \left( \| {\bm{\mathrm{r}}_{\mathcal{C}_{kl}}}\|^2_{{\bm{\Sigma}}_{\mathcal{C}_{kl}}} \right)
\label{baeqa}
\end{split}
\end{equation}
where $\bm{\mathrm{r}}_p$, $\bm{\mathrm{r}}_{\mathcal{I}_{k-1k}}$, $\bm{\mathrm{r}}_{\mathcal{C}_{kl}}$ are prior error, temporal IMU error and reprojection error respectively, as well as $\bm{\Sigma}_p$, ${\bm{\Sigma}}_{\mathcal{I}_{k-1k}}$, ${\bm{\Sigma}}_{\mathcal{C}_{kl}}$ are the corresponding covariance matrices, and $\rho$ is the Huber robust cost function. The optimization problem can be interpreted as a factor graph shown in Fig. \ref{ba}.

Therefore the best estimate for variable $\bm{\mathcal{X}}$ can be obtained by minimizing the objective function on manifold:
\begin{equation}
\bm{\mathcal{X}}^\ast = \mathop{\mathrm{argmin}} \limits_{\bm{\mathcal{X}} \in \mathcal{M}} f(\bm{\mathcal{X}})
\end{equation}

Detailed IMU and visual residuals are provided in the following subsections. The least squares problem are solved by Gauss-Newton method implemented in g2o\cite{G2O} or ceres solver.

\subsection{Inertial Measurement Model} \label{baimu}
IMU measurements arrive at a much higher frequency than the visual measurements. So in order to avoid the frequent integration whenever the linearization point changes, we adopt the IMU preintegration approach proposed in \cite{GTSAM}. The IMU preintegraton is independent of the initial conditions, and can incorporate the change of IMU biases. The concept was firstly proposed in \cite{VIA}.

We integrate all the IMU measurements $\{ \widetilde{\bm{\mathrm{a}}}_k, \ \widetilde{\bm{\mathrm{\omega}}}_k \}$ between keyframes i and j to compute the IMU preintegration $\Delta {\widetilde{\bm{\mathrm{I}}}_{ij}} = [ \Delta \widetilde{\bm{\mathrm{R}}}_{ij}, \Delta \widetilde{\bm{\mathrm{p}}}_{ij}, \Delta \widetilde{\bm{\mathrm{v}}}_{ij}]$ on manifold as:
\begin{equation}
\begin{split}
&\Delta \widetilde{\bm{\mathrm{R}}}_{ij} = \prod_{k=i}^{j-1} \mathrm{Exp}\left(( \widetilde{\bm{\mathrm{\omega}}}_{k}-\bm{\mathrm{b}}_{g_{i}}) \Delta t \right) \\
&\Delta \widetilde{\bm{\mathrm{p}}}_{ij} = \sum_{k=i}^{j-1} \left(\Delta \widetilde{\bm{\mathrm{v}}}_{ik} \Delta t + \frac{1}{2} \Delta \widetilde{\bm{\mathrm{R}}}_{ik} (\widetilde{\bm{\mathrm{a}}}_{k}-\bm{\mathrm{b}}_{a_{i}}) \Delta t^2 \right)\\
&\Delta \widetilde{\bm{\mathrm{v}}}_{ij} = \sum_{k=i}^{j-1} \Delta \widetilde{\bm{\mathrm{R}}}_{ik}(\widetilde{\bm{\mathrm{a}}}_{k}-\bm{\mathrm{b}}_{a_{i}}) \Delta t\\
\end{split}
\end{equation}


Furthermore, given a bias update $\delta \bm{\mathrm{b}}$ and using the first-order expansion, the preintegrated IMU measurement can be updated as:
\begin{equation}
\begin{split}
&\Delta \widetilde{\bm{\mathrm{R}}}_{ij}(\bm{\mathrm{b}}_{g_i}) = \Delta \widetilde{\bm{\mathrm{R}}}_{ij}(\bar{\bm{\mathrm{b}}}_{g_i}) \mathrm{Exp}\left( \frac{\partial \Delta \bar{\bm{\mathrm{R}}}_{ij}}{\partial \bm{\mathrm{b}}_g} \delta \bm{\mathrm{b}}_{g} \right)\\
&\Delta \widetilde{\bm{\mathrm{p}}}_{ij}(\bm{\mathrm{b}}_{g_i}, \bm{\mathrm{b}}_{a_i}) = \Delta \widetilde{\bm{\mathrm{p}}}_{ij}(\bar{\bm{\mathrm{b}}}_{g_i}, \bar{\bm{\mathrm{b}}}_{a_i}) + \frac{\partial \Delta \bar{\bm{\mathrm{p}}}_{ij}}{\partial \bm{\mathrm{b}}_g} \delta \bm{\mathrm{b}}_{g} + \frac{\partial \Delta \bar{\bm{\mathrm{p}}}_{ij}}{\partial \bm{\mathrm{b}}_a} \delta \bm{\mathrm{b}}_{a}\\
&\Delta \widetilde{\bm{\mathrm{v}}}_{ij}(\bm{\mathrm{b}}_{g_i}, \bm{\mathrm{b}}_{a_i}) = \Delta \widetilde{\bm{\mathrm{v}}}_{ij}(\bar{\bm{\mathrm{b}}}_{g_i}, \bar{\bm{\mathrm{b}}}_{a_i}) + \frac{\partial \Delta \bar{\bm{\mathrm{v}}}_{ij}}{\partial \bm{\mathrm{b}}_g} \delta \bm{\mathrm{b}}_{g} + \frac{\partial \Delta \bar{\bm{\mathrm{v}}}_{ij}}{\partial \bm{\mathrm{b}}_a} \delta \bm{\mathrm{b}}_{a} \\
\end{split}
\end{equation}
where Jacobians $\frac{\partial \Delta (\cdot)}{\partial \bm{\mathrm{b}}_{\cdot}}$ describe how a change in the bias estimate effects the preintegrated IMU measurements. The derivation of the Jacobians can be found in \cite{GTSAM}. From geometric constraints, we get the IMU measurement as:
\begin{equation}
\begin{split}
&\Delta \widetilde{\bm{\mathrm{R}}}_{ij}(\bm{\mathrm{b}}_{g_i}) = {\bm{\mathrm{R}}^{W}_{B_{i}}}^\mathrm{T} \bm{\mathrm{R}}^{W}_{B_{j}} \mathrm{Exp}(\delta \bm{\mathrm{\xi}}_{ij})\\
&\Delta \widetilde{\bm{\mathrm{p}}}_{ij}(\bm{\mathrm{b}}_{g_i}, \bm{\mathrm{b}}_{a_i}) = {\bm{\mathrm{R}}^{W}_{B_{i}}}^\mathrm{T} (\bm{\mathrm{p}}^{W}_{B_{j}} - \bm{\mathrm{p}}^{W}_{B_{i}} - \bm{\mathrm{v}}^{W}_{B_{i}} \Delta t_{ij} - \frac{1}{2} \bm{\mathrm{g}}^{W} \Delta t_{ij}^2) \\
& \ \ \ \ \ \ \ \ \ \ \ \ \ \ \ \ \ \ \ \ \ \ \ \ \ \ \ \ \ \ \ \ \ \ \ \ \ \ \ \ \ \ \ \  \ \ \ \ \ \ \ \ \ \ \ \ \ \ \ + \delta \bm{\mathrm{p}}_{ij} \\
&\Delta \widetilde{\bm{\mathrm{v}}}_{ij}(\bm{\mathrm{b}}_{g_i}, \bm{\mathrm{b}}_{a_i}) = {\bm{\mathrm{R}}^{W}_{B_{i}}}^\mathrm{T} (\bm{\mathrm{v}}^{W}_{B_{j}} - \bm{\mathrm{v}}^{W}_{B_{i}} - \bm{\mathrm{g}}^{W} \Delta t_{ij}) + \delta \bm{\mathrm{v}}_{ij}
\label{preintegrationmeasurementmodel}
\end{split}
\end{equation}
where ${[\delta \bm{\mathrm{\xi}}_{ij}^\mathrm{T} \ \delta \bm{\mathrm{t}}_{ij}^\mathrm{T} \ \delta \bm{\mathrm{v}}_{ij}^\mathrm{T}]}^\mathrm{T} \in \mathcal{N}(\bm{\mathrm{0}}, \bm{\Sigma}_{\mathcal{I}_{ij}})$ is zero-mean white Gaussian noise. Given the above measurement model, the IMU preintegration residual ${\bm{\mathrm{r}}_{\Delta_{ij}}} = {[\bm{\mathrm{r}}_{\Delta \bm{\mathrm{R}}_{ij}}^\mathrm{T} \ \bm{\mathrm{r}}_{\Delta \bm{\mathrm{p}}_{ij}}^\mathrm{T} \ \bm{\mathrm{r}}_{\Delta \bm{\mathrm{v}}_{ij}}^\mathrm{T}]}^\mathrm{T} \in \mathbb{R}^9$ is:
\begin{equation}
\begin{split}
&\bm{\mathrm{r}}_{\Delta \bm{\mathrm{R}}_{ij}} = \mathrm{Log}\left(  { \left( \Delta \widetilde{\bm{\mathrm{R}}}_{ij}(\bar{\bm{\mathrm{b}}}_{g_i}) \mathrm{Exp}\left( \frac{\partial \Delta \bar{\bm{\mathrm{R}}}_{ij}}{\partial \bm{\mathrm{b}}_g} \delta \bm{\mathrm{b}}_{g} \right) \right)}^\mathrm{T} {\bm{\mathrm{R}}^{W}_{B_{i}}}^\mathrm{T} \bm{\mathrm{R}}^{W}_{B_{j}} \right)\\
&\bm{\mathrm{r}}_{\Delta \bm{\mathrm{p}}_{ij}} = {\bm{\mathrm{R}}^{W}_{B_{i}}}^\mathrm{T} (\bm{\mathrm{p}}^{W}_{B_{j}} - \bm{\mathrm{p}}^{W}_{B_{i}} - \bm{\mathrm{v}}^{W}_{B_{i}} \Delta t_{ij} - \frac{1}{2} \bm{\mathrm{g}}^{W} \Delta t_{ij}^2) \\
&\ \ \ \ \ \ \ \ \ \ -\left[ \Delta \widetilde{\bm{\mathrm{p}}}_{ij}(\bar{\bm{\mathrm{b}}}_{g_i}, \bar{\bm{\mathrm{b}}}_{a_i}) + \frac{\partial \Delta \bar{\bm{\mathrm{p}}}_{ij}}{\partial \bm{\mathrm{b}}_g} \delta \bm{\mathrm{b}}_{g} + \frac{\partial \Delta \bar{\bm{\mathrm{p}}}_{ij}}{\partial \bm{\mathrm{b}}_a} \delta \bm{\mathrm{b}}_{a}\right] \\
&\bm{\mathrm{r}}_{\Delta \bm{\mathrm{v}}_{ij}} = {\bm{\mathrm{R}}^{W}_{B_{i}}}^\mathrm{T} (\bm{\mathrm{v}}^{W}_{B_{j}} - \bm{\mathrm{v}}^{W}_{B_{i}} - \bm{\mathrm{g}}^{W} \Delta t_{ij}) \\
&\ \ \ \ \ \ \ \ \ \ - \left[ \Delta \widetilde{\bm{\mathrm{v}}}_{ij}(\bar{\bm{\mathrm{b}}}_{g_i}, \bar{\bm{\mathrm{b}}}_{a_i}) + \frac{\partial \Delta \bar{\bm{\mathrm{v}}}_{ij}}{\partial \bm{\mathrm{b}}_g} \delta \bm{\mathrm{b}}_{g} + \frac{\partial \Delta \bar{\bm{\mathrm{v}}}_{ij}}{\partial \bm{\mathrm{b}}_a} \delta \bm{\mathrm{b}}_{a}\right]\\
\end{split}
\end{equation}

The corresponding covariance matrix ${\bm{\Sigma}}_{\Delta_{ij}}$ can be calculated by incrementally propagating the preintegration noise from keyframe $i$ to $j$. The detailed derivatives about Jaocbians and the uncertainty propagation on manifold can refer to paper \cite{GTSAM}.

\begin{figure*}[!t]
\centering
\includegraphics[width=6in]{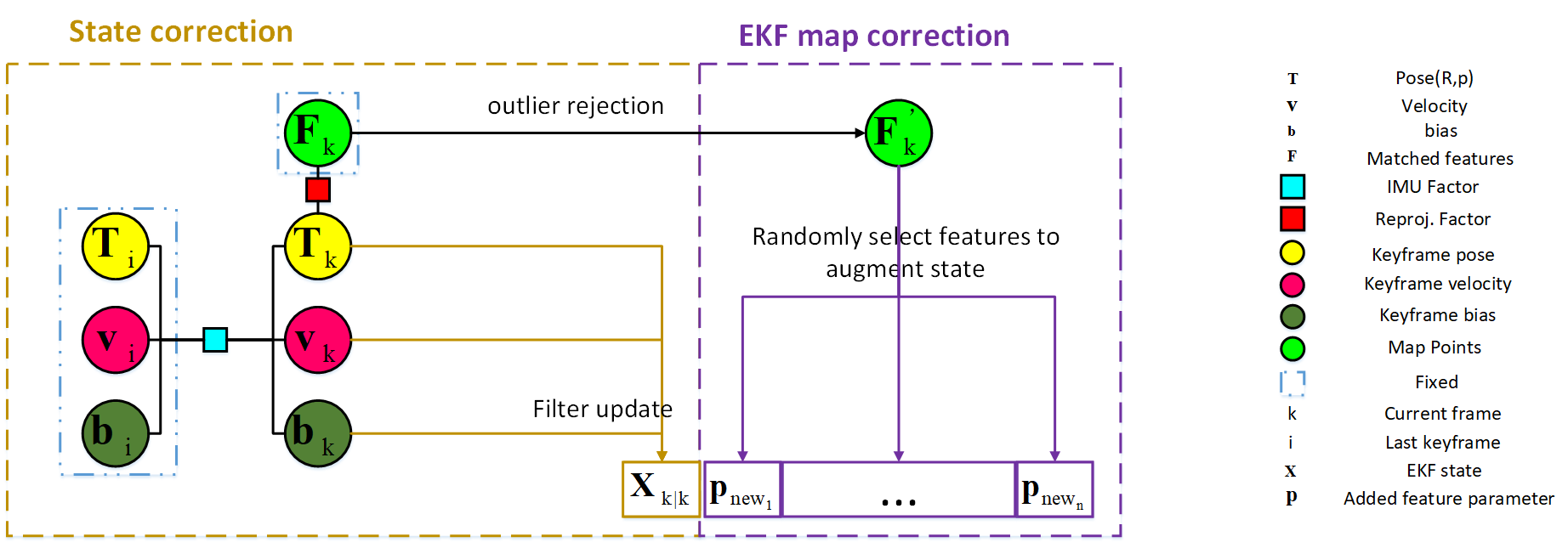}
\caption{A diagram shows how the feedback mechanism works for the current frame $k$ when the last keyframe $i$ just updated. The state of the current frame $k$ is firstly corrected by the state correction, and then new features are added to the state vector of the EKF by EKF map correction.}
\label{fig_feedback}
\end{figure*}

\subsection{Bias Model}\label{biasmodel}
Biases are slowly time-varying, so for the biases between consecutive keyframes i and j, we have:
\begin{equation}
    \bm{\mathrm{b}}_{g_j} = \bm{\mathrm{b}}_{g_i} + \bm{\mathrm{\eta}}_{b_{gd}}, \ \ \bm{\mathrm{b}}_{a_j} = \bm{\mathrm{b}}_{a_i} + \bm{\mathrm{\eta}}_{b_{ad}}
\end{equation}
where $\bm{\mathrm{\eta}}_{b_{gd}}$ and $\bm{\mathrm{\eta}}_{b_{ad}}$ are the discretized bias random walk with covariance $\bm{\Sigma}_{b_{gd}}$ and $\bm{\Sigma}_{b_{ad}}$. Therefore, we express the bias error ${\bm{\mathrm{r}}_{b} = [\bm{\mathrm{r}}_{g}^\mathrm{T} \ \bm{\mathrm{r}}_{a}^\mathrm{T}}] \in \mathbb{R}^6$ as:
\begin{equation}
\begin{split}
   \bm{\mathrm{r}}_{g} &=  \bm{\mathrm{b}}_{g_j} -\bm{\mathrm{b}}_{g_i}\\
   \bm{\mathrm{r}}_{a} &=  \bm{\mathrm{b}}_{a_j} -\bm{\mathrm{b}}_{a_i}
\end{split}
\end{equation}

\subsection{Visual Measurement Model} \label{bavisual}
Through the measurement model in \eqref{visualmm}, the reprojection residual $\bm{\mathrm{r}}_{\mathcal{C}_{kl}} \in \mathbb{R}^2$ for the $l^{th}$ map point seen by the $k^{th}$ keyframe is:
\begin{equation}
{\bm{\mathrm{r}}_{\mathcal{C}_{kl}}} = \pi\left( {\bm{\mathrm{R}}^B_C}^\mathrm{T} \left( {\bm{\mathrm{R}}^{W}_{B_{i}}}^\mathrm{T} (\bm{\mathrm{L}}_l^W - \bm{\mathrm{p}}^{W}_{B_{i}}) -\bm{\mathrm{p}}^B_C \right) \right) - \widetilde{\bm{\mathrm{z}}}_{kl}
\end{equation}

The corresponding covariance matrix ${{\bm{\Sigma}}_{\mathcal{C}_{kl}}}$ is equal to $\bm{\Sigma}_{\sigma_{kl}}$.

\subsection{Error Term Representation}
In this section, we give detailed representation for the IMU error term $\|{\bm{\mathrm{r}}_{\mathcal{I}_{ij}}}\|^2_{{\bm{\Sigma}}_{\mathcal{I}_{ij}}}$ and the reprojection error term $\|{\bm{\mathrm{r}}_{\mathcal{C}_{kl}}}\|^2_{{\bm{\Sigma}}_{\mathcal{C}_{kl}}}$ in \eqref{baeqa}. Given the inertial measurement model in Section \ref{baimu} and bias model in Section \ref{biasmodel}, the IMU error term is:
\begin{equation}
\|{\bm{\mathrm{r}}_{\mathcal{I}_{ij}}}\|^2_{{\bm{\Sigma}}_{\mathcal{I}_{ij}}} = \bm{\mathrm{r}}_{\Delta_{ij}}^\mathrm{T} {\bm{\Sigma}}_{\Delta_{ij}}^{-1} \bm{\mathrm{r}}_{\Delta_{ij}} + \bm{\mathrm{r}}_{b}^\mathrm{T} {\bm{\Sigma}}_{b_d}^{-1} \bm{\mathrm{r}}_{b}
\label{imuerrorterm}
\end{equation}
where ${\bm{\Sigma}}_{b_d} = \left[ \begin{array}{cc}
    \bm{\Sigma}_{b_{gd}} & \bm{\mathrm{0}} \\
    \bm{\mathrm{0}} & \bm{\Sigma}_{b_{ad}}
\end{array} \right]$.

In addition, given the visual measurement model in Section \ref{bavisual}, the reprojection error term is:
\begin{equation}
\|{\bm{\mathrm{r}}_{\mathcal{C}_{kl}}}\|^2_{{\bm{\Sigma}}_{\mathcal{C}_{kl}}} = \bm{\mathrm{r}}_{\mathcal{C}_{kl}}^\mathrm{T} \bm{\Sigma}_{\mathcal{C}_{kl}}^{-1} \bm{\mathrm{r}}_{\mathcal{C}_{kl}}
\label{visualerrorterm}
\end{equation}

\section{tightly-coupled monocular VISLAM}\label{sec:approach}
In this section, we introduce our tightly-coupled visual-inertial monocular SLAM approach, which combines the EKF-based VIO front-end with the BA and loop closure back-end to provide the accurate and robust state estimation. An overview of our system is shown in Fig.1. When new inertial sensor is used, we firstly perform the EKF VIO alone with initial bias value of zero to obtain a good initial bias estimates before starting the all system, then set it as the initial bias value of our system.

\subsection{Map Initialization}
The map initialization is in charge of constructing an initial set of map points for the subsequent nonlinear optimization and loop closure. The initial map is created according to the estimated state from the EKF VIO. Since the accuracy of the initially created map will pose big influence on the accuracy of the whole system, we create initial map after running the EKF system about 10 seconds to converge.

Firstly, we extract ORB features in the current frame $k$, and search for feature matches with the reference frame $r$. If sufficient feature correspondences are found, we perform the next step, else set the current frame as reference frame. The second step is to check the parallax of each correspondence, and pick out a set of feature matches $\mathcal{F}$ which have sufficient parallax. When the size of $\mathcal{F}$ is greater than a threshold, using the estimated pose from the EKF based VIO, we triangulate the matched features $\mathcal{F}$. Then, if enough map points are successfully created, a full BA combining reprojection error term and temporal IMU error term is applied to refine the initial map. Finally, using the optimized state and map to correct the EKF state according to the feedback mechanism described in Section \ref{sec:feedback}.

\subsection{FRONT-END}

\subsubsection{\textbf{State Estimation}}
We firstly perform the VIO described in Section \ref{sec:filter} to estimate the state of current frame $\bm{\mathrm{X}}_{B_{k}} = \{ {\bm{\mathrm{\xi}}^{W}_{B_k}} \ {\bm{\mathrm{p}}^{W}_{B_k}} \ {\bm{\mathrm{v}}^{W}_{B_k}} \ {\bm{\mathrm{b}}_{a_k}} \ {\bm{\mathrm{b}}_{g_k}}\}$.
In order to ensure the real-time performance, we only remain those features visible in current frame to limit the number of keypoints in the state vector. In our work, we maintain 50 features in the EKF state. Finally, if the map is updated in the back-end, the state of the current frame will be corrected according to the feedback mechanism described in Section \ref{sec:feedback}. In this way, the front-end can always provide reliable state estimates even if we run for long periods of time.

\subsubsection{\textbf{Keyframe Selection}}
After the state of a frame is estimated by the VIO, we adopt three criteria to determine whether this frame is a keyframe. (1) The time interval from last keyframe is beyond a certain threshold. This criteria ensures the accuracy of the system. Because the IMU just provide valuable information in short-time, if the time interval from the last keyframe is too long, the IMU constraint between two keyframes will become inaccurate. (2) The nonlinear optimization in back-end is finished. This criteria makes as many keyframes as possible to enhance the motion tracking accuracy. (3) The rotation angle from last keyframe is beyond a certain threshold. This criteria ensures the reconstruction of the globally consistent map.

\indent Once a frame is selected as a keyframe, we extract the ORB features for the new keyframe and trigger the back-end to make the pose estimation more accurate.

\subsection{BACK-END}
Once a new keyframe is inserted, the nonlinear optimization described in Section \ref{sec:optimization} is performed to optimization the local map in a parallel thread. After the local BA is finished, some redundant keyframes will be culled to make the factor graph more concise. In loop closure thread, place recognition is performed. Once a loop is detected, a $\bm{\mathrm{Sim}}(3)$ optimization and a full BA is performed to eliminate the accumulated drift. We refer the interested readers to papers \cite{ORBVIO}\cite{ORBSLAM} for more details.

\subsection{FEEDBACK MECHANISM}\label{sec:feedback}
The EKF-based VIO can estimate the frame state efficiently. However since the accumulation of the linearization errors and the absence of the loop closure, the error of the state estimate will accumulate as time goes on. If the state provided in the front-end drift too much, the local BA will be hard to find best estimates. Therefore in order to constrain the error of the state estimated from VIO, we provide following feedback mechanism. The feedback mechanism is invoked whenever the map in the back-end is updated and contains two steps:

\subsubsection{\textbf{State Correction}}
After the BA and loop closure are performed in the back-end, the state estimation of the last keyframe $i$ is accurate enough. The key observation of the state correction is to improve the state estimation of the current frame $k$ by leveraging the optimized state of the last keyframe $i$. Therefore, given the estimated state $\bm{\mathrm{X}}_{B_{k}}$ from VIO, we optimize the state of the current frame by performing the nonlinear optimization as shown in Fig. \ref{fig_feedback}, that is minimizing the following objective function:
\begin{equation}
\begin{array}{l}
\bm{\mathrm{X}}_{B_{k}}^\ast = \mathop {\arg \min }\limits_{\bm{\mathrm{X}}_{B_{k}}^\ast} \rho\left(\|{\bm{\mathrm{r}}_{\mathcal{I}_{ik}}}\|^2_{{\bm{\Sigma}}_{\mathcal{I}_{ik}}}\right) + \sum\limits_{l \in {F_k}} \rho \left( \| {\bm{\mathrm{r}}_{\mathcal{C}_{kl}}}\|^2_{{\bm{\Sigma}}_{\mathcal{C}_{kl}}} \right)
\end{array}
\end{equation}
where $F_k$ denotes the features matched with the map points in current frame $k$. The form of $\|{\bm{\mathrm{r}}_{\mathcal{I}_{ik}}}\|^2_{{\bm{\Sigma}}_{\mathcal{I}_{ik}}}$ and $\| {\bm{\mathrm{r}}_{\mathcal{C}_{kl}}}\|^2_{{\bm{\Sigma}}_{\mathcal{C}_{kl}}}$ is the same as \eqref{imuerrorterm} and \eqref{visualerrorterm} respectively. Then the optimized state $\bm{\mathrm{X}}_{B_{k}}^\ast$ and its covariance matrix $\bm{\mathrm{\Upsilon}}_{B_{k}}$ obtained from the optimization are used to update the EKF state of the current frame as:
\begin{equation}
\begin{split}
&\bm{\mathrm{H}}_k^\ast = \left[ \begin{array}{cc} \bm{\mathrm{I}}_{15 \times 15} &\bm{\mathrm{0}}_{15 \times 6m} \end{array} \right]\\
&\bm{\mathrm{r}}_k^\ast = \bm{\mathrm{X}}_{B_{k}}^\ast - \bm{\mathrm{X}}_{B_{k|k}} \\
&\bm{\mathrm{K}}_k^\ast = \bm{\mathrm{P}}_{k|k} {\bm{\mathrm{H}}_k^\ast}^T {(\bm{\mathrm{H}}_k^\ast \bm{\mathrm{P}}_{k|k}  {\bm{\mathrm{H}}_k^\ast}^T + \bm{\mathrm{\Upsilon}}_{B_{k}})}^{-1}\\
&\bm{\mathrm{X}}_{{k|k}}^\ast = \bm{\mathrm{X}}_{{k|k}} + \bm{\mathrm{K}}_k^\ast \bm{\mathrm{r}}_k^\ast\\
&\bm{\mathrm{P}}_{{k|k}}^\ast = (\bm{\mathrm{I}}_{15+6m} - \bm{\mathrm{K}}_k^\ast\bm{\mathrm{H}}_k^\ast) \bm{\mathrm{P}}_{{k|k}}
\label{feedbackupdate}
\end{split}
\end{equation}

\subsubsection{\textbf{EKF Map Correction}}
It is well known that the estimated pose relative to the map in the state vector is not drift. Therefore, as long as the position of map points in the EKF state vectorF is consistent with the optimized global map, the accuracy of the estimated state will accordingly increase. Therefore after the IMU state in EKF state vector is updated by state correction, we will add features in the optimized consistent map to the state vector. We denote $F_k^{'}$ as a set of features matched with the optimized map in current frame $k$, the outliers are removed based on the optimized state. If we need to add $n$ new features to the EKF state, we randomly select n features in $F_k^{'}$, then compute their initial position and add it to the filter state vector. For selected new feature $l$, the initial position ${\bm{\mathrm{p}}_l^W}^\ast = {[x_l^\ast \ y_l^\ast \ z_l^\ast \ \theta_l^\ast \ \phi_l^\ast \ \rho_l^\ast]}^T$ is set as follows. ${[x_l^\ast \ y_l^\ast \ z_l^\ast]}^T$ is computed as \eqref{stateaugmentationxyz} and ${[\theta_l^\ast \ \phi_l^\ast]}^T$ is computed as \eqref{stateaugmentationthetaphi}\eqref{stateaugmentationuv} using the updated state $\bm{\mathrm{X}}_{{k|k}}^\ast$ from \eqref{feedbackupdate} and the observation of the selected new features in the current frame. In addition, for computing $\rho_l^\ast$, we firstly transform the map point in world frame $\bm{\mathrm{L}}_l^W$ to the current camera frame $\bm{\mathrm{L}}_l^{C_k}$:
\begin{equation}
\bm{\mathrm{L}}_l^{C_k} = \left[\begin{array}{c} x_L \\ y_L \\ z_L \end{array} \right] = {({\bm{\mathrm{R}}^W_{B_k}}^\ast \bm{\mathrm{R}}^B_C)}^T \left( \bm{\mathrm{L}}_l^W - ({\bm{\mathrm{R}}^W_{B_k}}^\ast \bm{\mathrm{R}}^B_C + {\bm{\mathrm{p}}^W_{B_k}}^\ast) \right)
\end{equation}

Then the initial inverse depth is obtained by $\rho_l^\ast = \frac{1}{\| \bm{\mathrm{L}}_l^{C_k} \|}$, and its variance is set as follows:
\begin{equation}
\sigma_{\rho_l}^\ast = \bm{\mathrm{J}}_L * \bm{\Sigma}_L * \bm{\mathrm{J}}_L^T + \bm{\mathrm{J}}_{Rt} * \bm{\Sigma}_{Rt} * \bm{\mathrm{J}}_{Rt}^T
\end{equation}
where $\bm{\mathrm{J}}_L = -\frac{1}{\| \bm{\mathrm{L}}_l^{C_k} \|^3} {\bm{\mathrm{L}}_l^{C_k}}^T {({\bm{\mathrm{R}}^W_{B_k}}^\ast \bm{\mathrm{R}}^B_C)}^T$ and $\bm{\mathrm{J}}_{Rt} = -\frac{1}{\| \bm{\mathrm{L}}_l^{C_k} \|^3} {\bm{\mathrm{L}}_l^{C_k}}^T \left[ {\bm{\mathrm{R}}^B_C}^T \left( {\bm{\mathrm{R}}^W_{B_k}}^{\ast T} (\bm{\mathrm{L}}_l^W - {\bm{\mathrm{p}}^W_{B_k}}^\ast)\right)^\wedge \ - {\bm{\mathrm{R}}^B_C}^T \right]$ are the Jacobians of $\rho_l^\ast$ with respect to $\bm{\mathrm{L}}_l^W$ and $[{\bm{\mathrm{R}}^W_{B_k}}^\ast \ {\bm{\mathrm{p}}^W_{B_k}}^\ast]$ on manifold. Besides, $\bm{\Sigma}_L$ and $\bm{\Sigma}_{Rt}$ are the covariance matrix for $\bm{\mathrm{L}}_l^W$ and $[{\bm{\mathrm{R}}^W_{B_k}}^\ast \ {\bm{\mathrm{p}}^W_{B_k}}^\ast]$, which is computed from the BA in the back-end.

\section{EXPERIMENTS}\label{experiment}
We make a complete evaluation of the proposed algorithm qualitatively and quantitatively on the EuRoC dataset\cite{EUROC}. The dataset contains 11 data sequences, which was recorded from a flying MAV in two different 30$m^2$ indoor rooms and a $300m^2$ industrial environment. Depending on the illumination, texture and motion dynamics, the data sequences are classified as easy, medium and difficult levels. The dataset provides synchronized global shutter WVGA stereo images at 20Hz, IMU measurements at 200Hz and ground truth state at 200Hz. We only use images from the left camera. Firstly, we evaluate the proposed algorithm qualitatively and quantitatively on the EuRoC dataset to show the accuracy of our system. Then we compare our method with other state-of-the-art approaches on the EuRoC dataset. Finally, the performance of our algorithm is validated again by indoor real-world experiments using the sensor of Intel RealSense ZR300. The experiments are performed on a laptop with Intel Core i5 2.2GHz CPU and an 8GB RAM.

\begin{figure}[!h]
\centering
\includegraphics[width=3.5in]{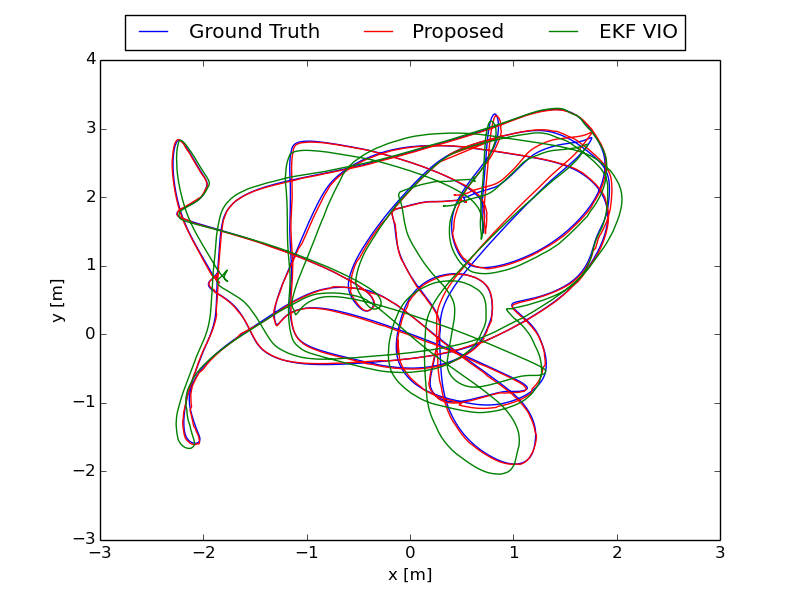}
\caption{Comparisons of the ground truth, the trajectories estimated by our algorithm and EKF VIO on V1\underline{\hspace{0.3em}}02\underline{\hspace{0.3em}}medium sequence, which is viewed from the gravity direction.}
\label{v102tra}
\end{figure}

\begin{figure}[!h]
\centering
\includegraphics[width=3.5in]{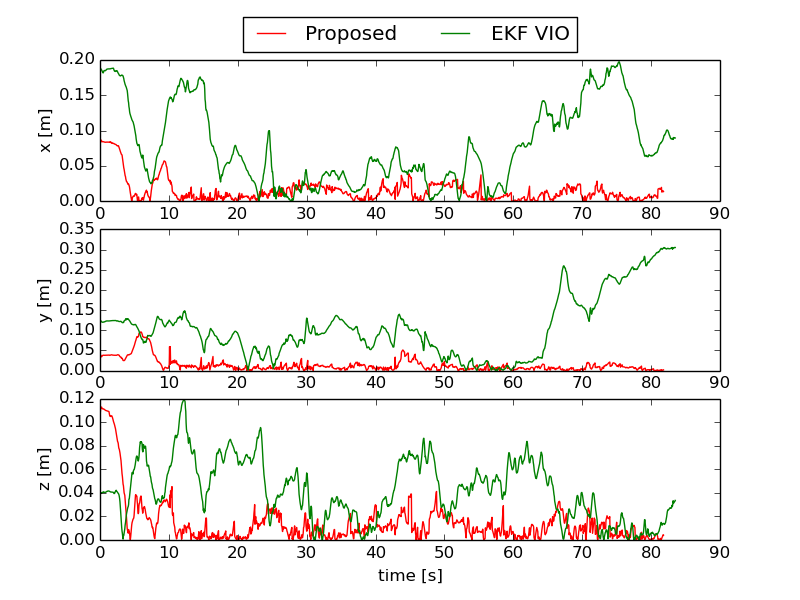}
\caption{Translation error of our algorithm and EKF VIO on V1\underline{\hspace{0.3em}}02\underline{\hspace{0.3em}}medium sequence.}
\label{v102error}
\end{figure}

\begin{figure}[!h]
\centering
\includegraphics[width=3.5in]{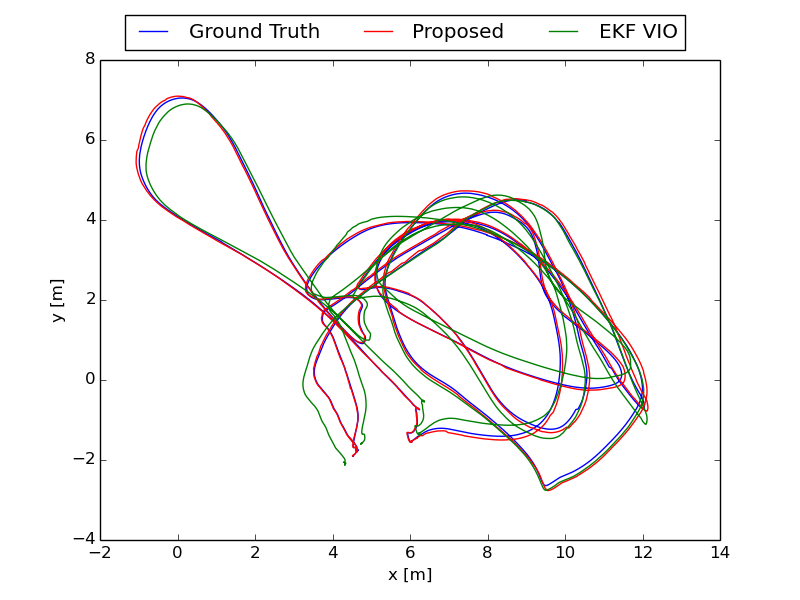}
\caption{Comparisons of the ground truth, the trajectories estimated by our algorithm and EKF VIO on MH\underline{\hspace{0.3em}}03\underline{\hspace{0.3em}}medium sequence, which is viewed from the gravity direction.}
\label{mh03tra}
\end{figure}

\begin{figure}[!h]
\centering
\includegraphics[width=3.5in]{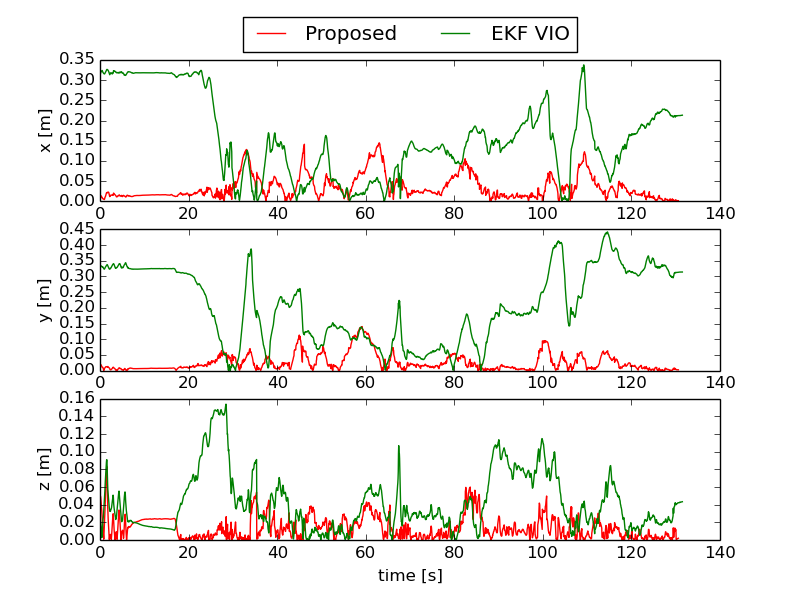}
\caption{Translation error of our algorithm and EKF VIO on MH\underline{\hspace{0.3em}}03\underline{\hspace{0.3em}}medium sequence.}
\label{mh03error}
\end{figure}

\subsection{Algorithm Evaluation}
We successfully perform our algorithm on all 11 sequences of EuRoC dataset in real-time. Fig.
\ref{v102tra} and Fig. \ref{mh03tra} show the comparisons of the ground truth, the trajectory estimated by our algorithm and EKF VIO on V1\underline{\hspace{0.3em}}02\underline{\hspace{0.3em}}medium and MH\underline{\hspace{0.3em}}02\underline{\hspace{0.3em}}easy sequence respectively. The corresponding x,y,z translation error versus time is shown in Fig. \ref{v102error} and Fig. \ref{mh03error}. The estimated trajectories are aligned with the ground truth using the method of Horn\cite{Horn}. As evident, the translation error of our approach is smaller than the error of EKF VIO, thus proving the superiority of our algorithm towards the EKF VIO method.

\begin{table}[!h]
\renewcommand{\arraystretch}{1.3}
\caption{Translation RMSE of the trajectories estimated from the proposed method and EKF VIO on the EuRoC MAV dataset}
\label{RMSE}
\centering
\begin{tabular}{|c|c|c|c|}
\hline
Sequence & \makecell[tl]{Our Method \\ With Loop} & \makecell[tl]{Our Method \\ Without Loop} & EKF\\
\hline
V1\underline{\hspace{0.3em}}01\underline{\hspace{0.3em}}easy & 0.080 & 0.080 & 0.087\\
\hline
V1\underline{\hspace{0.3em}}02\underline{\hspace{0.3em}}medium & 0.043 & 0.099 & 0.170\\
\hline
V1\underline{\hspace{0.3em}}03\underline{\hspace{0.3em}}difficult & 0.124 & 0.245 & 0.301\\
\hline
\hline
V2\underline{\hspace{0.3em}}01\underline{\hspace{0.3em}}easy & 0.052 & 0.052 & 0.082\\
\hline
V2\underline{\hspace{0.3em}}02\underline{\hspace{0.3em}}medium & 0.042 & 0.042 & 0.191\\
\hline
V2\underline{\hspace{0.3em}}03\underline{\hspace{0.3em}}difficult & 0.074 & 0.275 & 0.368\\
\hline
\hline
MH\underline{\hspace{0.3em}}01\underline{\hspace{0.3em}}easy & 0.021 & 0.021 & 0.175\\
\hline
MH\underline{\hspace{0.3em}}02\underline{\hspace{0.3em}}easy & 0.071 & 0.071 & 0.277\\
\hline
MH\underline{\hspace{0.3em}}03\underline{\hspace{0.3em}}medium & 0.061 & 0.061 & 0.307\\
\hline
MH\underline{\hspace{0.3em}}04\underline{\hspace{0.3em}}difficult & 0.064 & 0.064 & 0.309\\
\hline
MH\underline{\hspace{0.3em}}05\underline{\hspace{0.3em}}difficult & 0.048 & 0.056 & 0.529\\
\hline
\end{tabular}
\end{table}

For quantitative analysis, table \ref{RMSE} shows the translation Root Mean Square Error(RMSE) of the estimated trajectory for each sequence, as proposed in \cite{ATE}. The proposed method has achieved the average translation RMSE of 0.082m, 0.056m and 0.053m for V1, V2 and MH sequences with respect to 0.186m, 0.213m, 0.319m of EKF VIO system, which illustrates that our method reduced the error of 55\%, 73\%, 83\% for V1, V2 and MH sequences. From the third and fourth columns of the table \ref{RMSE}, we can know that adding BA and feedback mechanism to EKF VIO system, the RMSE of sequences are much reduced, this is since (1) BA is able to relinearize measurement models to properly deal with the nonlinearity of the system, (2) increasing the number of feature matches in the window of local BA can greatly improve the accuracy, and (3) feedback of the optimized state and map to VIO can improve the consistency and accuracy of the EKF system. However, the advantage of BA and feedback mechnism is not well demonstrated in V1\underline{\hspace{0.3em}}02\underline{\hspace{0.3em}}medium, V1\underline{\hspace{0.3em}}03\underline{\hspace{0.3em}}difficult and V2\underline{\hspace{0.3em}}03\underline{\hspace{0.3em}}difficult sequences, because in which fast rotation and low texture are frequently happened. Both fast rotation and low texture result in fewer feature correspondences, and thereby make less constraints in local BA, thus the accuracy of the system is not improved greatly. Therefore, in these sequences, adding a loop closure achieved much better accuracy by eliminating the accumulated error when revisiting an already mapped area.

While adding BA, loop closure and feedback mechanism improves accuracy of the system, it increases the computational cost of the system due to the need of (1) extracting a lot of features for BA and loop closure, and (2) pose optimization and EKF update of the feedback mechanism. The computational increase only occurs in selected keyframes, so the proposed method will not incur too much computational cost. Our algorithm requires approximately $27$ msec for processing each image, so it can still run in real time, at about 40Hz.

\begin{table*}
\caption{Translation RMSE of the trajectories estimated from different approaches on the EuRoC MAV dataset. The best results are given in bold.}
\label{RMSECOMPARISON}
\centering
\begin{tabular}{|c|c|c|c|c|c|c|c|}
\hline
sequence & \makecell[tl]{Our Method \\ With Loop} & \makecell[tl]{Our Method \\ Without Loop} & \makecell[tl]{VINS-MONO \\ With Loop} & \makecell[tl]{VINS-MONO \\ Without Loop} & ORB-VIN & OKVIS & ROVIO\\
\hline
V1\underline{\hspace{0.3em}}01\underline{\hspace{0.3em}}easy & 0.080 & 0.080 & 0.081 & 0.088 & $\bm{0.027}$ & 0.089 & 1.412\\
\hline
V1\underline{\hspace{0.3em}}02\underline{\hspace{0.3em}}medium & 0.043 & 0.099 & 0.042 & 0.068 & $\bm{0.028}$ & 0.141 & 0.160\\
\hline
V1\underline{\hspace{0.3em}}03\underline{\hspace{0.3em}}difficult & $\bm{0.124}$ & 0.245 & 0.156 & 0.160 & X & 0.262 & 0.170\\
\hline
\hline
V2\underline{\hspace{0.3em}}01\underline{\hspace{0.3em}}easy & 0.052 & 0.052 & 0.063 & 0.068 & $\bm{0.032}$ & 0.135 & 0.236\\
\hline
V2\underline{\hspace{0.3em}}02\underline{\hspace{0.3em}}medium & 0.042 & 0.042 & 0.066 & 0.084 & $\bm{0.041}$ & 0.155 & 0.408\\
\hline
V2\underline{\hspace{0.3em}}03\underline{\hspace{0.3em}}difficult & $\bm{0.074}$ & 0.275 & 0.157 & 0.159 & $\bm{0.074}$ & 0.279 & 0.213\\
\hline
\hline
MH\underline{\hspace{0.3em}}01\underline{\hspace{0.3em}}easy & $\bm{0.021}$ & $\bm{0.021}$ & 0.098377 & 0.301814 & 0.075 & 0.309 & 0.354\\
\hline
MH\underline{\hspace{0.3em}}02\underline{\hspace{0.3em}}easy & $\bm{0.071}$ & $\bm{0.071}$ & 0.152 & 0.249 & 0.084 & 0.293 & 0.594\\
\hline
MH\underline{\hspace{0.3em}}03\underline{\hspace{0.3em}}medium & $\bm{0.061}$ & $\bm{0.061}$ & 0.080 & 0.173 & 0.087 & 0.310 & 0.310\\
\hline
MH\underline{\hspace{0.3em}}04\underline{\hspace{0.3em}}difficult & $\bm{0.064}$ & $\bm{0.064}$ & 0.129 & 0.323 & 0.217 & 0.360 & 1.058\\
\hline
MH\underline{\hspace{0.3em}}05\underline{\hspace{0.3em}}difficult & $\bm{0.048}$ & 0.063 & 0.077 & 0.257 & 0.082 & 0.404 & 1.241\\
\hline
\end{tabular}
\end{table*}

\begin{figure}[!ht]
\centering
\includegraphics[width=3.5in]{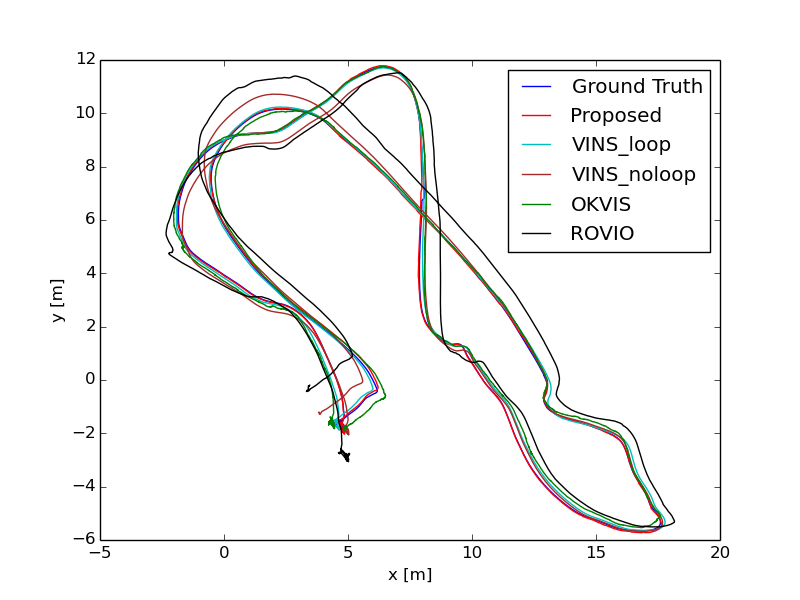}
\caption{Comparisons of the ground truth, the trajectories estimated by our algorithm and state-of-the-art methods on MH\underline{\hspace{0.3em}}04\underline{\hspace{0.3em}}difficult sequence, which is viewed from the gravity direction.}
\label{mh04tracomp}
\end{figure}

\begin{figure}[!ht]
\centering
\includegraphics[width=3.5in]{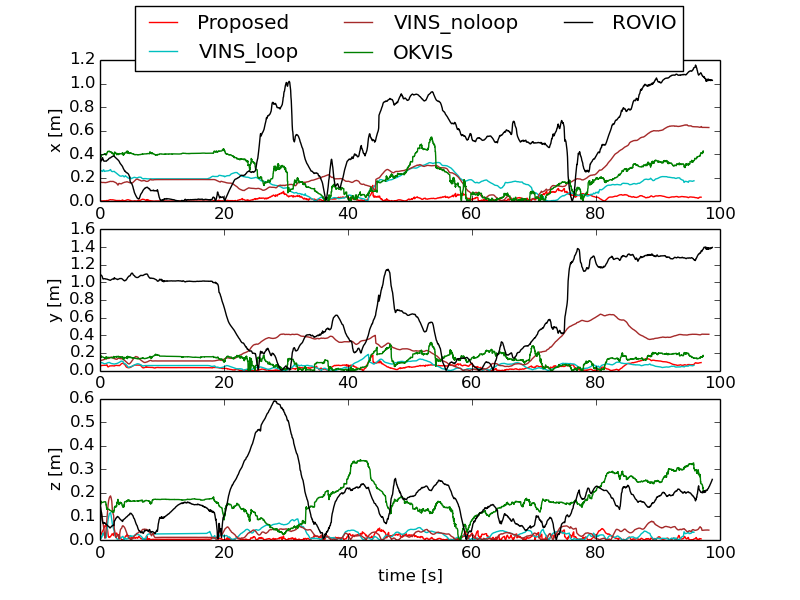}
\caption{Translation error of the estimated trajectories for MH\underline{\hspace{0.3em}}04\underline{\hspace{0.3em}}difficult sequence.}
\label{mh04errorcomp}
\end{figure}

\subsection{Comparison to State-of-the-art Algorithms}
We compare the proposed method with the state-of-the-art VINS-MONO \cite{VINSMONO1}\cite{VINSMONO2}, OKVIS \cite{OKVIS}, ROVIO \cite{ROBUSTDIRECT} and ORB-VISLAM\cite{ORBVIO} method. VINS-MONO, OKVIS and ROVIO are open-source and contain the default parameters for the EuRoC dataset, for fair comparison, we only use left image. ORB-VISLAM shows its results in EuRoC dataset, which allowing for a direct comparison.

A comparison of the translation RMSE of the estimated trajectories on EuRoC dataset are shown in table \ref{RMSECOMPARISON}, X means the concerned method fails to run in the sequence. From these results, we can draw following conclusions. ORB-VISLAM and our algorithm have obtained the best accuracy, this is because local BA in both methods is performed in a parallel thread, so more feature correspondences are used in local BA. Besides, both methods can close loop to eliminate the accumulated error. Howerver, ORB-VISLAM fails to track the V1\underline{\hspace{0.3em}}03\underline{\hspace{0.3em}}difficult sequence. In comparison, VINS-MONO and OKVIS perform BA in the thread of tracking, so the number of features contained in local BA must be limited to ensure the real-time performance, therefore leading to slightly worse accuracy. VINS-MONO with loop can achieve better accuracy than OKVIS, due to its capability to close loop. In addition, ROVIO is an EKF method and not able to close loop, its the linearization error and the error accumulation make the method obtain diminished localization error. However, since it is a direct method, it can achieve minor drift in fast motion sequences of V1\underline{\hspace{0.3em}}03\underline{\hspace{0.3em}}difficult and V2\underline{\hspace{0.3em}}03\underline{\hspace{0.3em}}difficult.

For sequence MH\underline{\hspace{0.3em}}04\underline{\hspace{0.3em}}difficult, the estimated trajectories are shown in Fig. \ref{mh04tracomp}, and translation errors versus time are shown in Fig. \ref{mh04errorcomp}. In this sequence, our loop closure is not triggered, however in the error plot, our method still achieved smallest translation error, which can prove the superiority of our algorithm again.

\begin{figure}[!ht]
\centering
\includegraphics[width=3.5in]{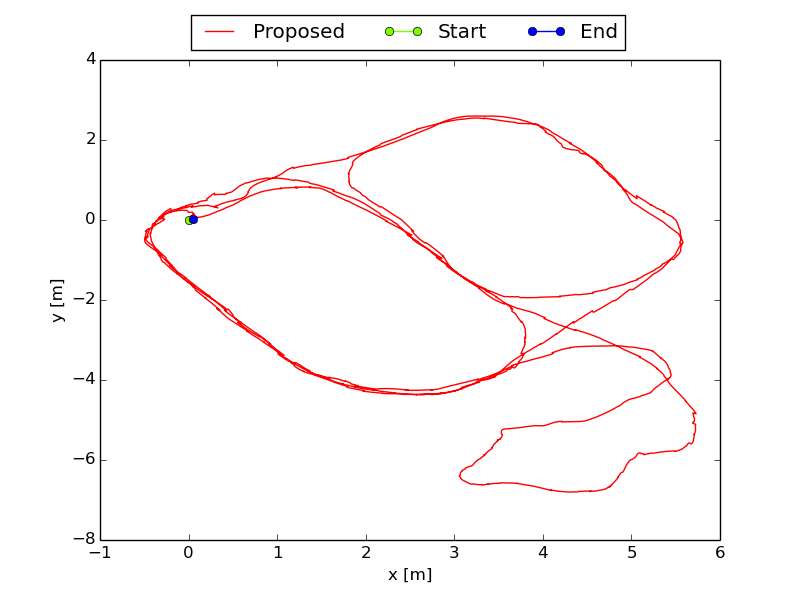}
\caption{The estimated trajectory of the indoor real-world experiment.}
\label{realsense}
\end{figure}

\subsection{Indoor Real-world Experiment}
We perform the indoor experiment in an 60$m^2$ office environment using the monocular-inertial Realsense ZR 300 sensor suite that provides images at the frequency of 20 Hz and IMU measurements at 200 Hz. As shown in the accompanying video{\footnote{\url{https://youtu.be/5_G8jUOjtN0}}}, we hold the sensor suite by hand and walk in normal pace in the office, and we starts and ends at the same location. Fig. \ref{realsense} shows the estimated trajectory, from which we can know there is no noticeable drifts occurred when we circle indoor. The end-to-end error is 0.055m with respect to the total length of 82m, it is just the 0.067\% of the total trajectory length.

\section{Conclusion and future work}\label{conclusion}
In this paper, we have presented a tightly-coupled monocular VISLAM system, which robustly tracks camera motion by EKF VIO, and perform non-linear optimization and loop closure to solve the linearization issues of the EKF system and eliminate the accumulated error. We also proposed a feedback mechanism to directly improve the consistency and accuracy of the EKF system. Therefore, our algorithm has achieved high accuracy, the performance of the proposed method is validated through experiments.

Point feature-based monocular VISLAM is prone to fail in poorly textured scenes or motion blurred images. Therefore in the future, we inted to deal with these specific situations for better accuracy and robustness. We also aim to build dense map to assist the understanding of the environment.

\ifCLASSOPTIONcompsoc
  \section*{Acknowledgments}
\else
  \section*{Acknowledgment}
\fi

\indent \indent This paper is supported by National Science Foundation of China[grant number 61375081]; a special fund project of Harbin science and technology innovation talents research [grant number RC2013XK010002].

\appendix
\subsection{Matrices $\bm{\mathrm{\Phi}}_{k}$ and $\bm{\mathrm{G}}_{k}$}
The Jacobian matrix $\bm{\mathrm{\Phi}}_{k}$ in \eqref{errorstatetrasition} is:
\begin{equation}
\bm{\mathrm{\Phi}}_{k} = \left[ \begin{array}{ccccc} \bm{\mathrm{\Phi}}_{\xi\xi} & \bm{\mathrm{0}}_{3 \times 3} & \bm{\mathrm{0}}_{3 \times 3} & \bm{\mathrm{0}}_{3 \times 3} & \bm{\mathrm{\Phi}}_{\xi b_g}\\
\bm{\mathrm{0}}_{3 \times 3} & \bm{\mathrm{I}}_{3 \times 3} & \bm{\mathrm{I}}_{3 \times 3} \Delta t & \bm{\mathrm{0}}_{3 \times 3} & \bm{\mathrm{0}}_{3 \times 3}\\
\bm{\mathrm{\Phi}}_{v \xi} & \bm{\mathrm{0}}_{3 \times 3} & \bm{\mathrm{I}}_{3 \times 3} & \bm{\mathrm{\Phi}}_{v b_a} & \bm{\mathrm{0}}_{3 \times 3}\\
\bm{\mathrm{0}}_{3 \times 3} & \bm{\mathrm{0}}_{3 \times 3} & \bm{\mathrm{0}}_{3 \times 3} & \bm{\mathrm{I}}_{3 \times 3} & \bm{\mathrm{0}}_{3 \times 3}\\
\bm{\mathrm{0}}_{3 \times 3} & \bm{\mathrm{0}}_{3 \times 3} & \bm{\mathrm{0}}_{3 \times 3} & \bm{\mathrm{0}}_{3 \times 3} & \bm{\mathrm{I}}_{3 \times 3} \\
\end{array}\right]
\end{equation}
where $\bm{\mathrm{I}}_{3 \times 3}$ is the $3 \times 3$ identity matrix, and
\begin{equation*}
\begin{split}
&\bm{\mathrm{\Phi}}_{\xi\xi} = \bm{\mathrm{J}}_r^{-1}({\bm{\mathrm{\xi}}^{W}_{B_{k|k-1}}})\mathrm{Exp}\left((\widetilde{\bm{\mathrm{\omega}}}_{k-1}-\bm{\mathrm{b}}_{g_{k-1}}-\bm{\mathrm{n}}_{gd}) \Delta t \right)^T \\
&\bm{\mathrm{\Phi}}_{\xi b_g} = -\bm{\mathrm{J}}_r^{-1}({\bm{\mathrm{\xi}}^{W}_{B_{k|k-1}}}) \bm{\mathrm{J}}_r\left((\widetilde{\bm{\mathrm{\omega}}}_{k-1}-\bm{\mathrm{b}}_{g_{k-1}}-\bm{\mathrm{n}}_{gd}) \Delta t \right) \Delta t\\
&\bm{\mathrm{\Phi}}_{v \xi} = -\bm{\mathrm{R}}^{W}_{B_{k-1}} {\left((\widetilde{\bm{\mathrm{a}}}_{k-1}-\bm{\mathrm{b}}_{a_{k-1}}-\bm{\mathrm{n}}_{ad}) \Delta t \right)}^{\wedge}\\
&\bm{\mathrm{\Phi}}_{v b_a} = -\bm{\mathrm{R}}^{W}_{B_{k-1}} \Delta t
\end{split}
\end{equation*}

In addition, the Jacobian matrix $\bm{\mathrm{G}}_{k}$ in \eqref{errorstatetrasition} is:
\begin{equation}
\bm{\mathrm{G}}_{k} = \left[ \begin{array}{cc} \bm{\mathrm{0}}_{3 \times 3} & \bm{\mathrm{\Phi}}_{\xi n_g}\\
\bm{\mathrm{0}}_{3 \times 3} & \bm{\mathrm{0}}_{3 \times 3}\\
\bm{\mathrm{\Phi}}_{v n_a} & \bm{\mathrm{0}}_{3 \times 3}\\
\bm{\mathrm{I}}_{3 \times 3} & \bm{\mathrm{0}}_{3 \times 3}\\
\bm{\mathrm{0}}_{3 \times 3} & \bm{\mathrm{I}}_{3 \times 3} \\
\end{array}\right]
\end{equation}
with $\bm{\mathrm{\Phi}}_{\xi n_g} = \bm{\mathrm{\Phi}}_{\xi b_g}$ and $\bm{\mathrm{\Phi}}_{v n_a} = \bm{\mathrm{\Phi}}_{v b_a}$.

\subsection{Measurement matrices}
The Jacobian of the measurement model with respect to the IMU state in \eqref{ekfmeasurementmodel} is represented as:
\begin{equation}
\bm{\mathrm{H}}_{B_{kl}} = \left[ -\bm{\mathrm{H}}_{h\xi} \ -\bm{\mathrm{H}}_{hp} \ \bm{\mathrm{0}}_{2 \times 9} \right]
\end{equation}
with:
\begin{equation*}
\begin{split}
&\bm{\mathrm{H}}_{h\xi} = \frac{\partial \bm{\mathrm{h}}_{kl}}{\partial \bm{\mathrm{f}}_l^{C_k}} {\bm{\mathrm{R}}^{B}_{C}}^T \cdot \\
&{\left({\bm{\mathrm{R}}^{W}_{B_{k|k-1}}}^T \left( \rho_l \left(\left[ \begin{array}{c} x_l \\ y_l \\ z_l \end{array}\right] - \bm{\mathrm{p}}^{W}_{B_{k|k-1}} \right) + \bm{\mathrm{m}}(\theta_l,\phi_l) \right) \right)}^{\wedge} \\
&\bm{\mathrm{H}}_{hp} = - \rho_l \frac{\partial \bm{\mathrm{h}}_{kl}}{\partial \bm{\mathrm{f}}_l^{C_k}}\bm{\mathrm{R}}_{W}^{C_{k|k-1}}\\
&\frac{\partial \bm{\mathrm{h}}_{kl}}{\partial \bm{\mathrm{f}}_l^{C_k}} = \left[ \begin{array}{ccc} \frac{f_x}{z_l^{C_k}} & 0 & -\frac{f_x x_l^{C_k}}{z_l^{C_k2}} \\ 0 & \frac{f_y}{z_l^{C_k}} & -\frac{f_y y_l^{C_k}}{z_l^{C_k2}} \end{array}\right]
\end{split}
\end{equation*}
where $\bm{\mathrm{f}}_l^{C_k} = {[x_l^{C_k} \ y_l^{C_k} \ z_l^{C_k}]}^T$. In addition, the Jacobian of the measurement model with respect to the $l^{th}$ feature position in \eqref{ekfmeasurementmodel} is:
\begin{equation}
\bm{\mathrm{H}}_{f_{kl}} = \left[ -\bm{\mathrm{H}}_{hxyz} \ -\bm{\mathrm{H}}_{h\theta\phi} \ -\bm{\mathrm{H}}_{h\rho} \right]
\end{equation}
with:
\begin{equation*}
\begin{split}
&\bm{\mathrm{H}}_{hxyz} = \rho_l \frac{\partial \bm{\mathrm{h}}_{kl}}{\partial \bm{\mathrm{f}}_l^{C_k}} \bm{\mathrm{R}}_{W}^{C_{k|k-1}} \\
&\bm{\mathrm{H}}_{h\theta\phi} = \frac{\partial \bm{\mathrm{h}}_{kl}}{\partial \bm{\mathrm{f}}_l^{C_k}} \bm{\mathrm{R}}_{W}^{C_{k|k-1}} \left[ \begin{array}{cc} cos\phi_l cos\theta_l & -sin\phi_l sin\theta_l \\ 0 & -cos\phi_l \\ -cos\phi_l sin\theta_l & -sin\phi_l cos\theta_l \end{array}\right] \\
&\bm{\mathrm{H}}_{h\rho} = \frac{\partial \bm{\mathrm{h}}_{kl}}{\partial \bm{\mathrm{f}}_l^{C_k}} \bm{\mathrm{R}}_{W}^{C_{k|k-1}} \left( \left[ \begin{array}{c} x_l \\ y_l \\ z_l \end{array}\right] - \bm{\mathrm{p}}^{W}_{C_{k|k-1}} \right)
\end{split}
\end{equation*}

\subsection{State Augmentation Jacobian}
The Jacobian matrix used to augment the covariance matrix of the state vector is:
\begin{equation}
\bm{\mathrm{J}} = \left[ \begin{array}{cc} \bm{\mathrm{I}}_{15+6m} & \bm{\mathrm{0}}_{(15+6m) \times 6} \\ \bm{\mathrm{J}}_{X} &  \bm{\mathrm{J}}_{h\rho} \end{array}\right]
\end{equation}
with:
\begin{equation}
\bm{\mathrm{J}}_{X} = \left[ \begin{array}{cccc} - \bm{\mathrm{R}}^{W}_{B_{k|k}} {\bm{\mathrm{p}}^{B}_{C}}^{\wedge} & \bm{\mathrm{I}}_3 & \bm{\mathrm{0}}_{3 \times 9} & \bm{\mathrm{0}}_{3 \times 6m}\\
\begin{array}{c} \frac{\partial \breve{\theta_l} \breve{\phi_l}}{ \partial \bm{\mathrm{\xi}}_{B_{k|k}}}  \\  \bm{\mathrm{0}}_{1 \times 3} \end{array} &
\bm{\mathrm{0}}_{3 \times 3} & \bm{\mathrm{0}}_{3 \times 9} & \bm{\mathrm{0}}_{3 \times 6m} \end{array} \right]
\end{equation}
where:
\begin{equation*}
\begin{split}
&\frac{\partial \breve{\theta_l} \breve{\phi_l}}{ \partial \bm{\mathrm{\xi}}_{B_{k|k}}} = -\frac{\partial \breve{\theta_l} \breve{\phi_l}}{\partial \bm{\mathrm{\tau}}_{{lk}}^W} \bm{\mathrm{R}}^{W}_{B_{k|k}} {\left(\bm{\mathrm{R}}^{B}_{C} \left[ \begin{array}{c} \frac{u_{lk}-c_x}{f_x} \\ \frac{v_{lk}-c_y}{f_y} \\ 1 \end{array}\right] \right)}^\wedge\\
&\frac{\partial \breve{\theta_l} \breve{\phi_l}}{\partial \bm{\mathrm{\tau}}_{{lk}}^W} = \left[ \begin{array}{ccc} \frac{z_{lk}^W}{\zeta} & 0 & -\frac{x_{lk}^W}{\zeta} \\ \frac{x_{lk}^W y_{lk}^W}{\varsigma} & -\frac{{x_{lk}^W}^2+{z_{lk}^W}^2}{\varsigma} & \frac{y_{lk}^W z_{lk}^W}{\varsigma} \end{array}\right] \\
&\zeta = {x_{lk}^W}^2+{z_{lk}^W}^2\\
&\varsigma = ({x_{lk}^W}^2+{y_{lk}^W}^2+{z_{lk}^W}^2) \sqrt{{x_{lk}^W}^2+{z_{lk}^W}^2}
\end{split}
\end{equation*}
and
\begin{equation}
\bm{\mathrm{J}}_{h\rho} = \left[ \begin{array}{ccc}  \bm{\mathrm{0}}_{3 \times 2} & \bm{\mathrm{0}}_{3 \times 1}\\
\frac{\partial \breve{\theta_l} \breve{\phi_l}}{\partial \bm{\mathrm{\tau}}_{{lk}}^W}{\bm{\mathrm{R}}^{W}_{B_{k|k}}}\bm{\mathrm{R}}^{B}_{C} \left[ \begin{array}{cc} \frac{1}{f_x} & 0 \\ 0 & \frac{1}{f_y} \\ 0 & 0 \end{array} \right] & \bm{\mathrm{0}}_{2 \times 1} \\ 0 & 1 \end{array} \right]
\end{equation}





%
\bibliography{bibfile}
\bibliographystyle{unsrt}

\end{document}